\documentclass[10pt,twocolumn,letterpaper]{article}

\usepackage[pagenumbers]{cvpr} 

\usepackage{graphicx}
\usepackage{amsmath}
\usepackage{amssymb}
\usepackage{booktabs}

\usepackage[pagebackref,breaklinks,colorlinks]{hyperref}

\usepackage{multirow}
\usepackage{algorithm}
\usepackage{algorithmic}

\usepackage[capitalize]{cleveref}
\crefname{section}{Sec.}{Secs.}
\Crefname{section}{Section}{Sections}
\Crefname{table}{Table}{Tables}
\crefname{table}{Tab.}{Tabs.}

\begin{document}

\title{V2C: Visual Voice Cloning}

\author{Qi Chen$^{1,3}$~~~Yuanqing Li$^{2,3}$~~~Yuankai Qi$^{1}$~~~Jiaqiu Zhou$^{2,3}$~~~Mingkui Tan$^{2}$\footnotemark[2]~~~Qi Wu$^{1}$\footnotemark[2]\\
	$^1$University of Adelaide~~~$^2$South China University of Technology~~~$^3$Pazhou Lab\\
	{\tt\small \{qi.chen04, qi.wu01\}@adelaide.edu.au, qykshr@gmail.com} \\ 
	{\tt\small \{auyqli, mingkuitan\}@scut.edu.cn, mszjq@mail.scut.edu.cn}}


\maketitle

\begin{abstract}
Existing Voice Cloning (VC) tasks aim to convert a paragraph text to a speech with desired voice specified by a reference audio.
This has significantly boosted the development of artificial speech applications.
However, there also exist many scenarios that cannot be well reflected by these VC tasks, such as movie dubbing, which requires the speech to be with emotions consistent with the movie plots.
To fill this gap, in this work we propose a new task named Visual Voice Cloning (V2C), which seeks to convert a paragraph of text to a speech with both desired voice specified by a reference audio and desired emotion specified by a reference video.
To facilitate research in this field, we construct a dataset, V2C-Animation, and propose a strong baseline based on existing state-of-the-art (SoTA) VC techniques.
Our dataset contains 10,217 animated movie clips covering a large variety of genres (\eg, Comedy, Fantasy) and emotions (\eg, happy, sad).
%
We further design a set of evaluation metrics, named MCD-DTW-SL, which help evaluate the similarity between  ground-truth speeches and the synthesised ones.
%
Extensive experimental results show that even SoTA VC methods cannot generate satisfying speeches for our V2C task.
We hope the proposed new task together with the constructed dataset and evaluation metric will facilitate the research in the field of voice cloning and broader vision-and-language community.

\end{abstract}

\renewcommand{\thefootnote}{\fnsymbol{footnote}}
\footnotetext[2]{Corresponding authors.}
\renewcommand{\thefootnote}{\arabic{footnote}}

\section{Introduction}\label{sec:intro}




\begin{figure}[t]
    \centering
    \includegraphics[width=1.0\linewidth]{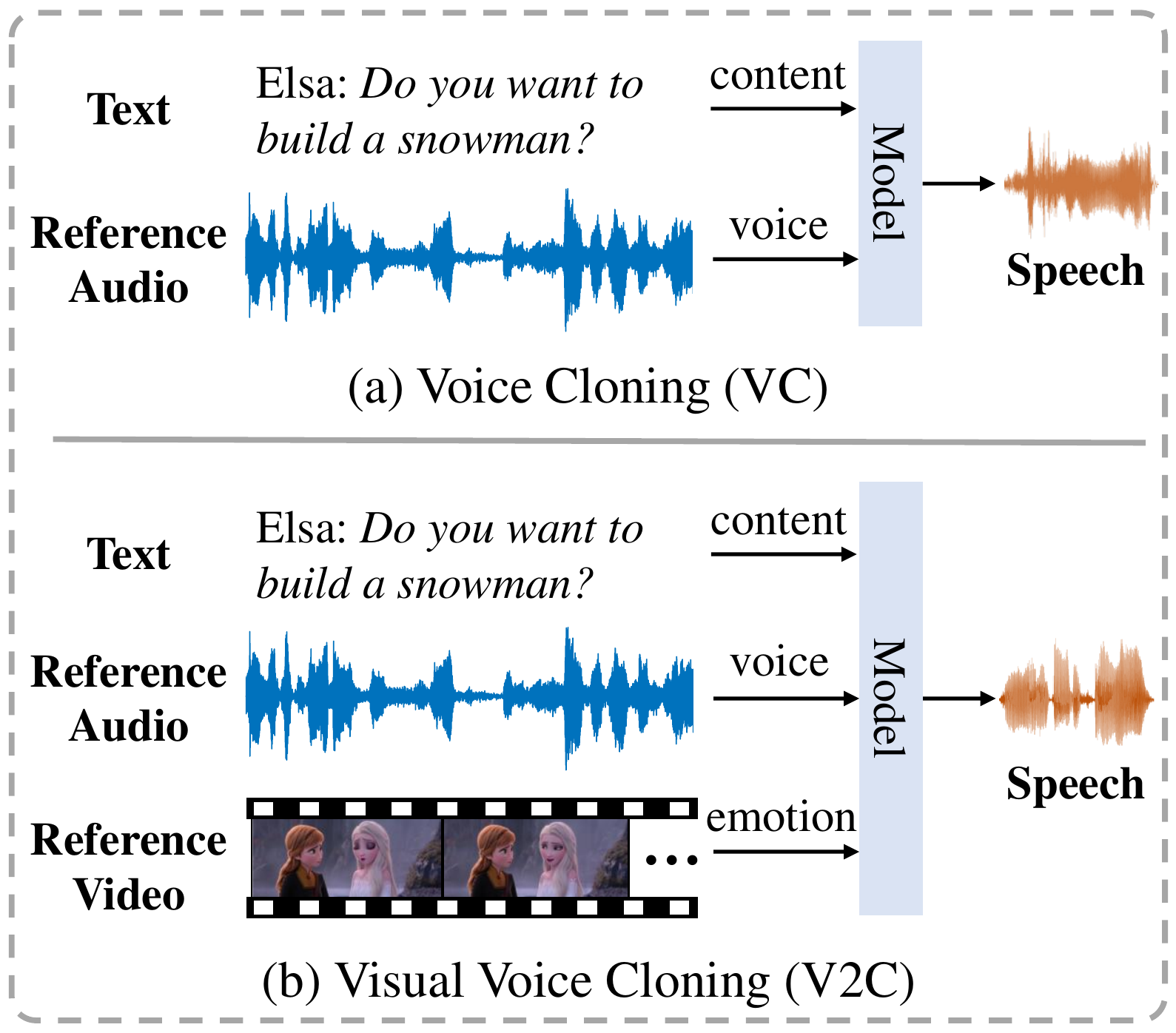}
    \caption{(a) Voice Cloning (VC) vs. (b) Visual Voice Cloning (V2C). Given an input triplet (\ie, subtitle/text, reference audio, and target video), our V2C task seeks to convert the text into a speech, which should be with the voice of reference audio and the emotion derived from reference video. Note that the reference audio only provides an expected voice while its content is irrelevant.}
    \label{fig:task}
\end{figure}

Voice Cloning (VC)~\cite{Arik2018NeuralVC, Chen2019SampleEA, Jia2018TransferLF, nachmani2018fitting}
aims to convert a paragraph of text to speech with the desired voice from a reference audio.
However, there exist many applications in the real world that require the generated speeches not only using a template voice but also being with rich emotions (\eg, angry, happy, and sad), such as movie dubbing. 
This is beyond the scope of conventional VC tasks (Figure~\ref{fig:task}(a)), as no extra guiding information can be used to generate desired tones and rhythms.
%
Considering that we humans accomplish the movie dubbing task with the most reference from visual observations (\eg, watching the movie to grasp the emotion of characters), we propose an  extension task of VC, namely Visual Voice Cloning (V2C).
%

An example of the proposed V2C task is shown in Figure~\ref{fig:task}(b). 
Unlike the conventional VC task, which converts text to speech only aided by a reference audio, our V2C task takes a triplet (text/subtitle, reference audio, reference video) as input and expects a resulting speech with the same voice but varying emotions derived from the reference video. The text/subtitle is the content that the generated speech needs to cover. The reference audio includes a pre-recorded voice of the target speaker from a different clip. And we aim to generate a speech with the voice in the reference audio and the character's visual emotion from the reference video, speaking the content in the given text.




The new task poses several novel challenges.
%
\textbf{First}, the conventional Voice Cloning (VC) methods~\cite{Arik2018NeuralVC, blaauw2019data, Chen2019SampleEA, Jia2018TransferLF, nachmani2018fitting} cannot well solve the V2C task as they focus only on how to convert the input text to speech with the voice/tone exhibited in the reference audio, without considering the emotion and context of the new speech.
However, in our V2C task (\eg, movie dubbing) the voice emotion is crucial for generating human-like speech.
\textbf{Second}, in our V2C task, the voice emotion should be derived from the reference video rather than the reference audio from an irrelevant clip. Taking movie dubbing as an example, it requires humans to grasp the emotions of characters by watching the corresponding movie clips and observing their performances (\eg, facial expressions or actions). Although several improved VC methods~\cite{skerry2018towards,wang2018style} also try to inject the voice emotion into their generated speech, they capture both emotion and voice from the reference audio, which cannot satisfy the requirements of V2C task.
In our V2C task, an ideal method should be able to disentangle voice and emotion from the reference audios and the reference videos, respectively.

As there is no off-the-shelf dataset suitable for the V2C task, we collect the first V2C-Animation dataset to facilitate the research in this field. 
%
The V2C-Animation dataset comprises 10,217 video clips with audios and subtitles, covering 26 animated movies with 153 characters (\ie, speakers) in total.
Our V2C dataset covers three modalities (\ie, text, audio and video) unlike the existing text-to-speech datasets~\cite{ljspeech17,panayotov2015librispeech,yamagishi2019cstr,zen2019libritts} or movie description datasets~\cite{rohrbach2015dataset,tapaswi2016movieqa}, which only focus on text and audio, or text and video.
Besides, we also provide emotion annotation (\eg, happy or sad) for each audio and video clip like~\cite{goodfellow2013challenges}.
To alleviate the impact from background music, we only extract the sound channel of the centre speaker, which mainly focuses on the sound of the speaking character. In this way, we ensure that all the audio clips only contain the sound from speaking characters.

To address the above challenges of V2C task, based on the widely used Text-to-Speech (TTS) framework (\ie, FastSpeech2~\cite{ren2020fastspeech}), we propose a new method called Visual Voice Cloning Network (V2C-Net), considering the emotion information derived from the reference video frames.
Moreover, based on MCD~\cite{kubichek1993mel}, we design an evaluation metric, called MCD-DTW weighted by Speech Length (MCD-DTW-SL), seeking to evaluate the generated speech effectively and automatically.

In summary, our contribution include:
\begin{itemize}
\vspace{-2mm}
    \item We propose a new task, namely Visual Voice Cloning (V2C). Given a triplet (\ie, text/subtitle, reference audio and reference video), the task seeks to convert the text into a speech with voice and emotion derived from reference audio and reference video, respectively.
    \vspace{-2mm}
    \item We collect the first V2C-Animation dataset, which consists of 26 animated movies, 153 characters, 10,217 video clips with the aligned audios and subtitles, covering three modalities (\ie, text, audio, video) and speakers' emotion.
    \vspace{-2mm}
    \item We design a new method, called Visual Voice Cloning Network (V2C-Net). Besides, to evaluate the generated speech automatically, we provide an advanced automatic evaluation metric, named MCD-DTW-SL.
\end{itemize}



\section{Related Work}
As the V2C is a new task, here we briefly review several closely related works in the fields of Text to Speech, Voice Cloning, and Prosody Transfer.

\noindent\textbf{Text to Speech.}
Many text-to-speech (TTS) synthesis methods~\cite{arik2017deep,kalchbrenner2018efficient,Wang2017TacotronTE,li2019neural,chen2021adaspeech,yan2021adaspeech} have been proposed to generate natural speech from text. 
%
Then, based on WaveNet, Deep Voice~\cite{arik2017deep} divides a TTS model into several modules, which are optimised independently. 
Wang~\etal~\cite{Wang2017TacotronTE} propose a new framework Tacotron, which integrates all the necessary stages in text-to-speech synthesis and
enables that the speech synthesis model can be optimised in an end-to-end manner.
Recently, TransformerTTS~\cite{li2019neural} introduces the structure of transformer~\cite{vaswani2017attention} into TTS task while Ren~\etal~\cite{ren2019fastspeech} propose a more efficient transformer (\ie, FastSpeech) by using non auto-regressive generation method. Based on FastSpeech, they further design an improved FastSpeech2~\cite{ren2020fastspeech}, which seeks to control the generated speech via the adjustment of pitch and energy.
%
However, the TTS task mainly focuses on how to convert natural language text to speech in a correct pronounce. Instead, our V2C task requires the generated speech to be additionally with a suitable voice emotion and tone.

\noindent\textbf{Voice Cloning.}
Unlike the TTS method, which synthesises speech only with a single voice, voice cloning (VC) task \cite{ Gibiansky2017DeepV2, Ping2018DeepV3,  Taigman2018VoiceLoopVF} seeks to generate multiple speeches with different voices. Based on Deep Voice~\cite{arik2017deep} and Tacotron~\cite{Wang2017TacotronTE}, Deep Voice 2 \cite{Gibiansky2017DeepV2} map the voices from different speakers into a common space and use the low-dimensional embedding from this common space as a condition to aid the generation process.
Jia \etal \cite{Jia2018TransferLF} propose a multi-speaker TTS framework, which consists of three sub-modules (\ie, encoder, synthesizer and vocoder), which is able to synthesise a high-quality speech from given text.
More recent extensions~\cite{Arik2018NeuralVC, blaauw2019data, Chen2019SampleEA, nachmani2018fitting} focus on synthesising the voice of an unseen person using a few samples only. Specifically, 
to synthesise a person's voice from only a few audio samples, Arik \etal \cite{Arik2018NeuralVC} study two approaches: speaker adaption and speaker encoding. The speaker adaption seeks to fine-tune a trained multi-speaker model for an unseen speaker using a few audio-text pairs while the speak encoding tries to directly estimate the speaker embedding from audios of an unseen speaker.
Chen \etal \cite{Chen2019SampleEA} propose an adaptive TTS system by using the meta-learning approach.
Different from VC task, our V2C additionally requires the prosody/tone of generated speeches to match with the reference video.

\noindent\textbf{Prosody Transfer.}
To produce realistic speech, 
prosody transfer (PT) \cite{bian2019multi, Hsu2019HierarchicalGM, skerry2018towards, stanton2018predicting, Valle2020MellotronME, wang2018style, Whitehill2020MultiReferenceNT} seeks to grasp prosody from reference audios. Specifically, extending from Tacotron~\cite{Wang2017TacotronTE}, Skerry-Ryan \etal \cite{skerry2018towards} propose an encoder architecture to learn a representation of prosody from reference spectrogram slices, which are derived from the reference audio.
Global Style Tokens (GSTs) \cite{wang2018style} models the styles of different speakers using an interpretable embedding, which can be used as a condition when transferring the different speaking styles. Based on the variational autoencoder (VAE) framework~\cite{Kingma2014AutoEncodingVB}, Hsu \etal \cite{Hsu2019HierarchicalGM} design a neural sequence-to-sequence TTS model, which categorises the speaking styles into several latent attributes and hence controls the speaking style via adapting these attributes. 
To transfer speaking style which is underrepresented in the dataset, Whitehill \etal \cite{Whitehill2020MultiReferenceNT} propose an adversarial cycle-consistent training procedure for a multi-reference neural TTS system. 
Overall, the goal of prosody transfer is to capture both the voice and emotion from the reference audio, 
and therefore the task is defined without the use of the information from visual side.
By contrast, V2C task is proposed to infer the voice emotion from a reference video, which has many real-world applications such as movie dubbing.

\section{V2C Task and V2C-Animation Dataset}

\subsection{Problem Definition for V2C Task}

Given a triplet $\mathcal{Z}=\{Z_{text}, Z_{audio}, Z_{video}\}$ (\ie, text, reference audio and reference video), our Visual Voice Cloning (V2C) task aims to generate a speech $\mathcal{Y}_{wave}$ (\ie, a waveform in time-domain) from text $Z_{text}$, which should use the voice of reference audio $Z_{audio}$ and be with the emotion derived from reference video $Z_{video}$.
In Figure~\ref{fig:task}, we take movie dubbing as an example. Given a movie clip (\ie, reference video), a subtitle (\ie, text) and a reference audio, we seek to synthesise a speech from subtitle according to both the character's emotion derived from movie and the voice from reference audio.


\subsection{Dataset Construction for V2C Task}\label{sec:data_collection}


The dataset for V2C task should cover all three modalities and the samples from different modalities need to be aligned with each other. As there is no an off-the-shelf dataset suitable for this new task, we collect the first V2C dataset, called V2C-Animation.

\begin{figure}[t]
    \centering
    \includegraphics[width=0.90\linewidth]{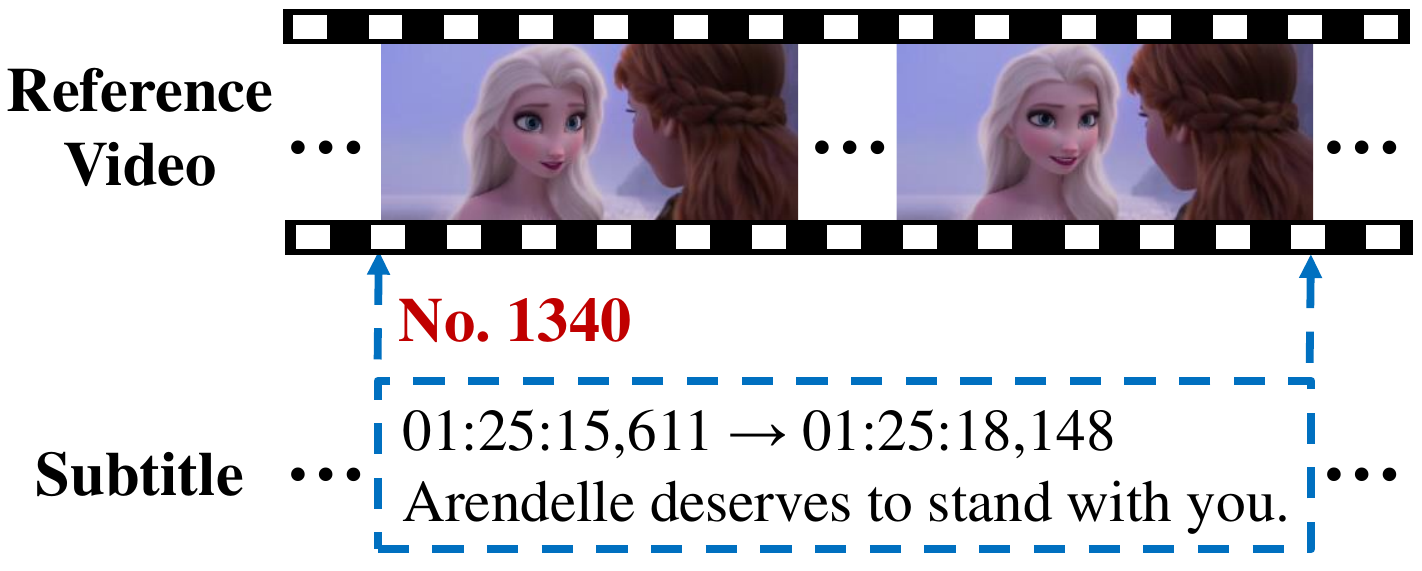}
    \vspace{-5pt}
    \caption{An example of how to cut a movie into a series of video clips according to SRT subtitle files. Note that the SRT files contain both starting and ending time-stamps for each video clip. No.~1340 refers to the sequential number of the current utterance.}
    \label{fig:examples_subtitle_crop_video}
    \vspace{-5pt}
\end{figure}

\noindent\textbf{Data Collection.}
We search for Blu-ray animated movies with the corresponding subtitles and then select a set of 26 movies of diverse genres.
Specifically, we first cut the movies into a series of video clips according to the subtitle files.
Here, we use an SRT type subtitle file. In addition to subtitles/texts, the SRT file also contains starting and ending time-stamps to ensure the subtitles match with video and audio, and sequential number of subtitles (\eg, No.~1340 in Figure~\ref{fig:examples_subtitle_crop_video}), which indicates the index of each video clip.
%
Based on the SRT file, we cut the movie into a series of video clips using the FFmpeg toolkit~\cite{tomar2006converting} (an automatic audio and video processing toolkit) and then extract the audio from each video clip by FFmpeg as well.
%
Note that, to alleviate the impact from the background music,
we only extract the sound channel of the centre speaker, which mainly focuses on the sound of the speaking character. 

\begin{table*}[t]
  \centering
  \resizebox{1.0\linewidth}{!}
  {
    \begin{tabular}{c|ccc|cc|cccccc}
    \toprule
    Dataset & Text  & Audio & Video & Identity & Emotion & \#Movies & \#Video Clips & \#Audio Clips & \#Speakers & Avg. S & Avg. A/V (s)  \\
    \midrule
    LJ Speech~\cite{ljspeech17} & $\surd$     & $\surd$     &  &  $\surd$ &    &   -    &    -   & 13100 & 1     & 17.23 & 6.57  \\
    LibriSpeech~\cite{panayotov2015librispeech} & $\surd$     & $\surd$     &   &  $\surd$ &     &   -    &   -    & 250698 & 2484  & 32.55 & 14.10  \\
    VCTK~\cite{yamagishi2019cstr} & $\surd$     & $\surd$     &   &  $\surd$ &     &   -    &   -    & 44070 & 108   & 7.41  & 3.59  \\
    LibriTTS~\cite{zen2019libritts} & $\surd$     & $\surd$     &   &  $\surd$ &     &    -   &   -    & 375086 & 2456  & 16.86 & 5.62  \\
    \midrule
    MPII-MD~\cite{rohrbach2015dataset} & $\surd$     &    &   $\surd$ &   &    & 94    & 68337 &    -   &    -   &    -   & 3.88  \\
    MovieQA~\cite{tapaswi2016movieqa} & $\surd$     &    &     $\surd$  &   &   & 140   & 6771  &    -   &   -    &   6.20    & 202.67  \\
    \midrule
    LRS2-main~\cite{chung2017lip} & $\surd$ & $\surd$     & $\surd$   &  &  &  -    & 48164 & 48164 & -   &  7.13  & 1$\sim$2 \\
    LRS2-pretrain~\cite{chung2017lip} & $\surd$ & $\surd$     & $\surd$  &  &  &  -    & 96318 & 96318 &  -  & 21.43  & $\sim$10 \\
    \midrule
    V2C-Animation & $\surd$ & $\surd$     & $\surd$     & $\surd$ &  $\surd$ &  26    & 10217 & 10217 & 153   & 6.51  & 2.40  \\
    \bottomrule
    \end{tabular}%
    }
  \caption{We compare our V2C-Animation dataset with several existing multi-modal datasets. ``Identity'' and ``Emotion'' indicate whether the datasets contain annotations corresponding to the speaker's identity and emotion. The notations ``\#Movies'', ``\#Video Clips'', ``\#Audio Clips'' and ``\#Speakers'' refer to the number of movies, videos, audios and speakers/characters, respectively. ``Avg.~S'' indicates the average length of subtitle while ``Avg.~A/V'' is the average duration of audio/video.
  }
  \label{tab:comparison_with_other_dataset}%
\end{table*}%

\noindent\textbf{Data Annotation and Organisation.}
Inspired by the organisation of LibriSpeech dataset~\cite{panayotov2015librispeech}, we categorise the obtained video clips, audios and subtitles into their corresponding characters (\ie, speakers) via a crowd-sourced service. To ensure that the characters appeared in the video clips are the same as the speaking ones, we manually remove the data example that does not satisfy the requirement.
Then, following the categories of FER-2013~\cite{goodfellow2013challenges} (a dataset for human facial expression recognition), we divide the collected video/audio clips into 8 types including angry, happy, sad, \etc. 
In this way, we totally collect a dataset with 10,217 video clips with paired audios and subtitles. 
All of the annotations, time-stamps of the mined movie clips and a tool to extract the triplet data will be released. 
%
We randomly split $60\%$ samples as training data, $10\%$ samples as validation data and $30\%$ samples as testing data. 
%

\subsection{V2C-Animation Dataset vs. Related Datasets}

We compare our V2C-Animation dataset against ones of the VC, TTS, and PT tasks.
%
In addition, we consider some Movie Description (MD) datasets and Lip Reading Sentence (LRS) datasets, which contain both video and text like ours.
Specifically, the VC/TTS/PT datasets include LJ Speech~\cite{ljspeech17}, LibriSpeech~\cite{panayotov2015librispeech}, VCTK~\cite{yamagishi2019cstr} and LibriTTS~\cite{zen2019libritts}, the MD datasets involve MPII-MD~\cite{rohrbach2015dataset} and MovieQA~\cite{tapaswi2016movieqa}, while LRS datasets contain LRS2~\cite{chung2017lip}\footnote{The LRS2 dataset has two subsets: LRS2-main and LRS2-pretrain. On LRS2-pretrain, the utterance of each video may contain multiple sentences. But each video only corresponds to a single sentence on LRS2-main. There is some overlap between LRS2-pretrain and LRS2-main sets.}.
From Table~\ref{tab:comparison_with_other_dataset}, the statistic results demonstrate that our V2C-Animation dataset is unique, covering all the three modalities (\ie, text, audio and video) with both identity and emotion annotations, while most others only focus on the two modalities and all of them are without emotion annotations. More dataset statistics can be found in the supplementary.

\begin{figure}[t]
    \centering
    \includegraphics[width=1.0\linewidth]{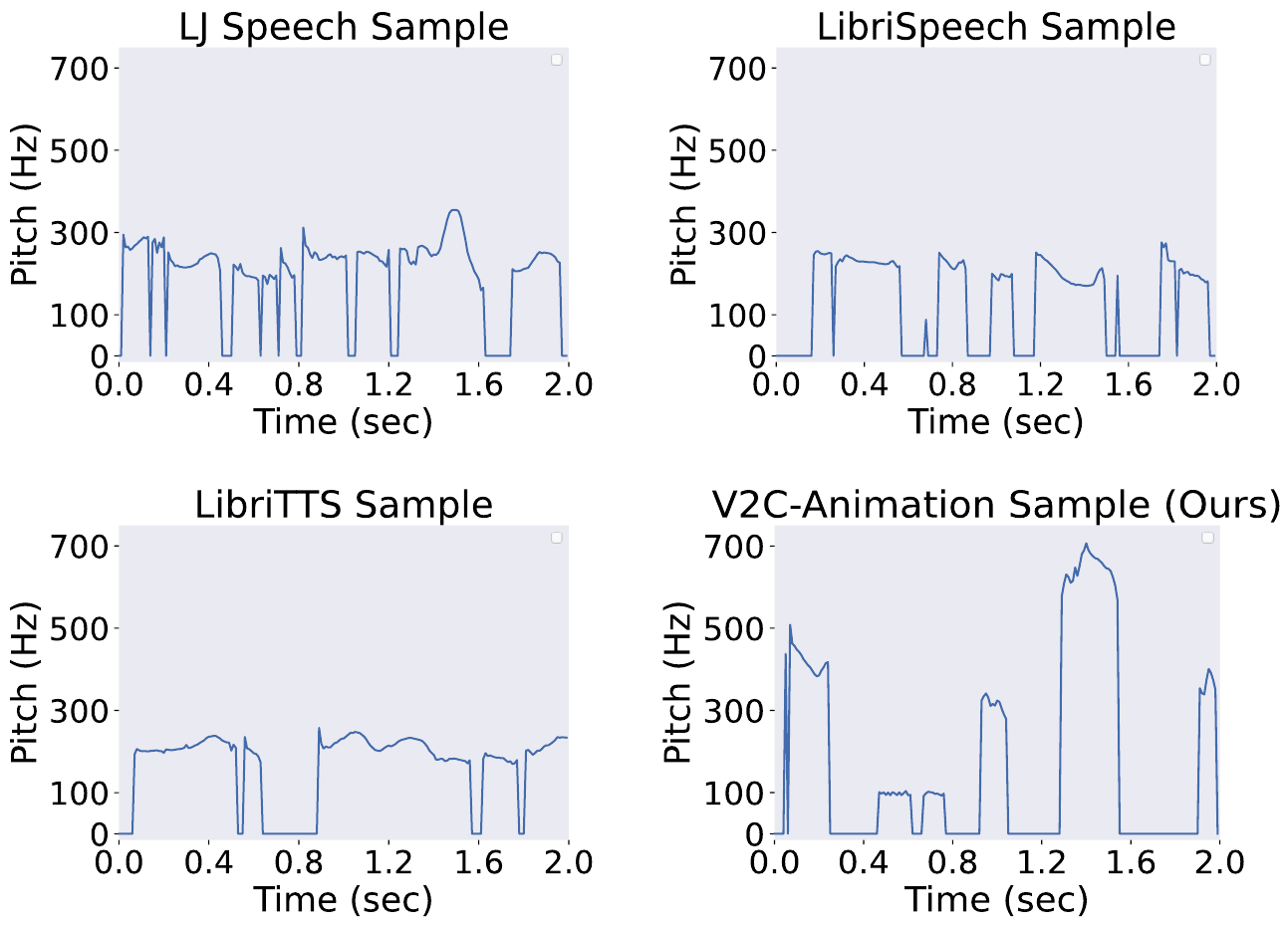}
    \caption{Examples from the existing Text-to-Speech (TTS) datasets (\ie, LJ speech, LibriTTS and LibriSpeech) and our V2C-Animation dataset. A pitch of 0 Hz refers to an unvoiced segment.}
    \label{fig:dataset_pitch_comparison}
\end{figure}

To further compare the differences between our V2C-Animation dataset and related datasets, following~\cite{skerry2018towards}, we visualise the pitch tracks of the samples from our dataset and others.
Specifically, we randomly select an audio sample from LJ Speech, LibriSpeech and LibriTTS, respectively.
Due to the varying lengths of audios,
for a fair comparison, we crop two seconds of audio from each compared sample. As shown in Figure~\ref{fig:dataset_pitch_comparison}, the audio pitches from the existing datasets are more smooth and the ranges of frequency (Hz) are narrower than ours.
%
Moreover, we provide average and variance values of the pitch tricks. Table~\ref{tab:average_variance_pitch_value} shows that the variance of our V2C-Animation is largest, which further demonstrates that our proposed dataset has a wider range of frequency (Hz).
Both the visual and statistical results demonstrate that our V2C-Animation dataset is more challenging due to the varied prosody.
%


\begin{table}[t]
  \centering
    \resizebox{0.92\linewidth}{!}
    {
    \begin{tabular}{c|cc}
    \toprule
    Dataset & LJ Speech & LibriSpeech \\
    \midrule
    Avg. P (Hz) & 127.27 $\pm$ 11800.96 & 88.15 $\pm$ 7313.39  \\
    \midrule
    \midrule
    Dataset & LibriTTS & V2C-Animation (Ours) \\
    \midrule
    Avg. P (Hz) & 93.97 $\pm$ 9295.67 & 117.99 $\pm$ 16910.77 \\
    \bottomrule
    \end{tabular}%
    }
    \caption{We compare the average and variance of pitch from our V2C-Animation dataset and the related datasets. ``Avg. P'' refers to the average values of pitches with the corresponding variances.}
  \label{tab:average_variance_pitch_value}%
\end{table}%

\section{Visual Voice Cloning Network (V2C-Net)}

\begin{figure*}[t]
    \centering
    \includegraphics[width=0.92\linewidth]{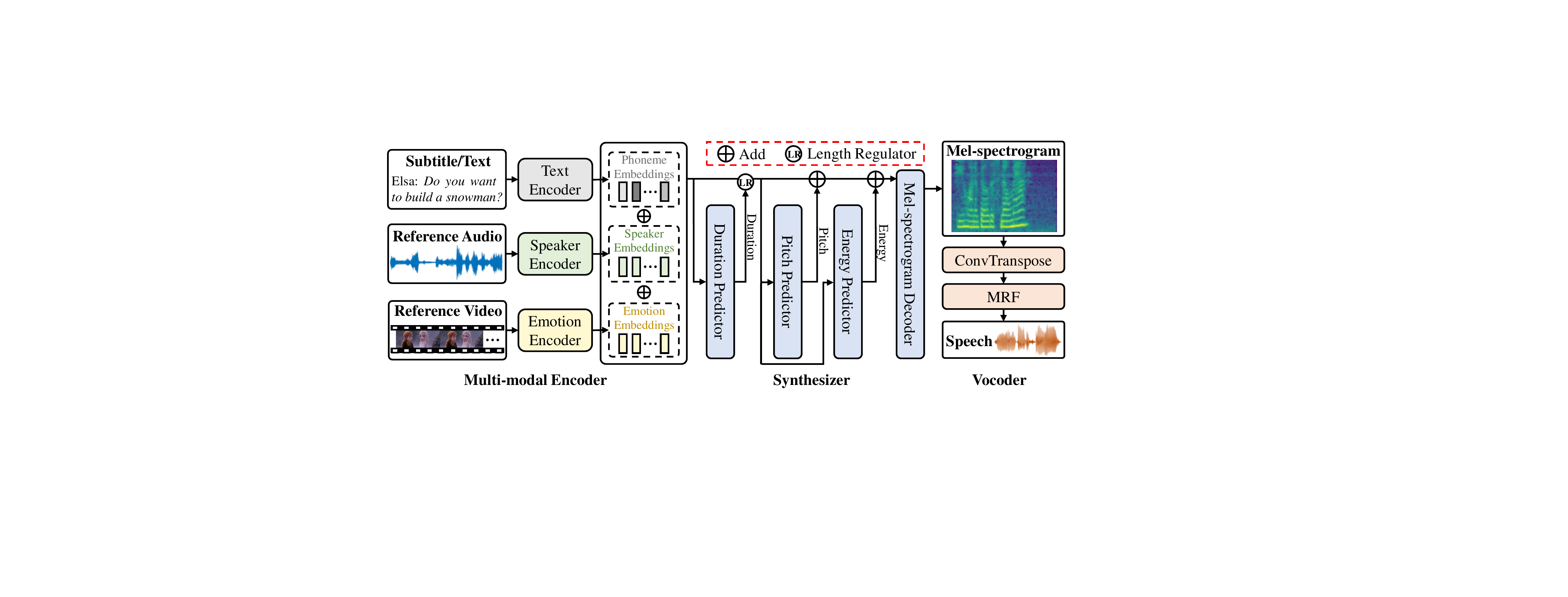}
    \vspace{-5pt}
    \caption{Overview of V2C-Net. It consists of three main components: a multi-modal encoder, a synthesizer and a vocoder. 
    A triplet (\ie, text, reference audio and reference video) is fed into the multi-modal encoder (Sec.~\ref{sec:encoder}) and it outputs three types of embeddings. 
    Based on these embeddings, the synthesizer (Sec.~\ref{sec:synthesizer}) generates a mel-spectrogram. 
    Finally, the mel-spectrogram is converted into the waveform (\ie, speech) by the vocoder (Sec.~\ref{sec:vocoder}).}
    \label{fig:overall_architecture}
\end{figure*}

For the V2C task, we propose a baseline model, called Visual Voice Cloning Network (V2C-Net), which is based on a widely used TTS framework FastSpeech2~\cite{ren2020fastspeech}. 
As shown in Figure~\ref{fig:overall_architecture}, our model contains three main components: a multi-modal encoder, a synthesizer and a vocoder. 
We input a triplet (\ie, text, reference audio, and reference video) into the encoder and output three types of features (\ie, phoneme, speaker, and emotion). 
Based on these features, we use the synthesizer to generate a mel-spectrogram (see Figure~\ref{fig:overall_architecture} right side), which is a time-frequency representation of the audio signal.
Last, we convert the generated mel-spectrogram into waveform (\ie, speech, see Figure~\ref{fig:overall_architecture} right-bottom) by the vocoder.

\subsection{Multi-modal Encoder for Feature Extraction}
\label{sec:encoder}

Given a triplet $\mathcal{Z}=\{Z_{text}, Z_{audio}, Z_{video}\}$, the output feature/embedding $\mathcal{X}$ from the multi-modal encoder is
\begin{equation}
    \begin{aligned}
        \mathcal{X} = \{\mathbf{x}_1, ..., \mathbf{x}_L\} = f(\mathcal{Z}),
     \end{aligned}
\end{equation}
where $L$ is the length of input sentence (\ie, the number of phoneme). $f(\cdot)$ is the multi-modal encoder, mainly containing three sub-modules: a text encoder $f_{txt}$, a speaker encoder $f_{spk}$, and an emotion encoder $f_{emo}$. Here, $Z_{text}$, $Z_{audio}$ and $Z_{video}$ indicate the input text, reference audio and reference video, respectively.
We obtain the $i$-th output feature $\mathbf{x}_i=\mathbf{o}_i\oplus\mathbf{u}\oplus\mathbf{v}$, where $\mathbf{o}_i$ is the embedding of the $i$-th phoneme derived from text $Z_{text}$. The embeddings $\mathbf{u}$ and $\mathbf{v}$ are the outputs of $f_{spk}$ and $f_{emo}$, respectively. The notation $\oplus$ indicates the operation of element-wise add. 


\noindent\textbf{Text encoder.}
Following the structure of FastSpeech2, we take 4 Feed-Forward Transformer (FFT) blocks~\cite{ren2019fastspeech} as our text encoder.
Based on such a text encoder, we produce a series of phoneme embeddings $\mathcal{O}=\{\mathbf{o}_1, ..., \mathbf{o}_L\}$ from an input text $Z_{text}$. Mathematically, the process can be defined as $\mathcal{O} = f_{txt} (Z_{text})$.

\noindent\textbf{Speaker encoder.}
To explore the voice characteristics of different speakers, we adopt a speaker encoder $f_{spk}$, which has the same architecture as~\cite{wan2018generalized}, comprising 3 LSTM layers and a linear layer. The speaker encoder first converts a sequence of mel-spectrogram frames, derived from the reference audio, to a series of hidden embeddings by LSTM, and then maps the last hidden embedding to a fixed-dimensional vector via the linear layer.
For convenience, we define the process as $\mathbf{u} = f_{spk}(\sigma(Z_{audio}))$, where $\mathbf{u}$ refers to the speaker embedding while $\sigma$ is a mapping function, converting the audio from waveform to mel-spectrogram.


\noindent\textbf{Emotion encoder.}
To exploit the emotion from video, we design an emotion encoder $E_E$, which captures the embedding of the whole video clip $Z_{video}$. To be specific, we use the I3D model~\cite{carreira2017quo} as our emotion encoder $f_{emo}$ and calculate the emotion embedding $\mathbf{v}$ by $\mathbf{v}=f_{emo}(Z_{video})$.

\subsection{Synthesizer for Mel-Spectrogram Generation}
\label{sec:synthesizer}

To generate the mel-spectrogram from the conditional phoneme embedding $\mathcal{X}=\{ \mathbf{x}_1, ..., \mathbf{x}_L\}$, we introduce a synthesizer inspired by FastSpeech2~\cite{ren2020fastspeech}, and obtain the predicted mel-spectrogram frames $\mathcal{Y}=\{ \mathbf{y}_1, ..., \mathbf{y}_T\}$,
where $T$ is the number of mel-spectrogram frames.
Here, the synthesizer mainly contains four parts: a duration predictor, a pitch predictor, an energy predictor, and a mel-spectrogram decoder.
Loss function of the synthesizer is
\begin{equation}
    \begin{aligned}
        \mathcal{L}_{S} = \mathcal{L}_{mel} + \lambda_1 \mathcal{L}_{dur} + \lambda_2 \mathcal{L}_{pitch} + \lambda_3 \mathcal{L}_{energy},
     \end{aligned}
     \label{eq:synthesizer_loss}
\end{equation}
where $\mathcal{L}_{mel}$, $\mathcal{L}_{dur}$, $\mathcal{L}_{pitch}$ and $\mathcal{L}_{energy}$ refer to the losses of mel-spectrogram, duration predictor, pitch predictor and energy predictor, respectively. The $\lambda_1$, $\lambda_2$ and $\lambda_3$ are hyper-parameters, and we set $\lambda_1=\lambda_2=\lambda_3=1$ in practice. The details are depicted in the following.

\noindent\textbf{Duration Predictor.}
To alleviate the problem of length mismatch between the input embedding $\mathcal{X}$ and mel-spectrogram frames $\mathcal{Y}$ (\ie, $L\neq T$), we introduce a duration predictor $S_d$, which takes the embedding $\mathcal{X}$ as input and predicts the duration $\mathcal{D}=\{d_1, ..., d_i ..., d_L\} = \{S_d(\mathbf{x}_1), ..., S_d(\mathbf{x}_i) ..., S_d(\mathbf{x}_L)\}$ of each phoneme embedding.
The $i$-th phoneme duration $d_i$ indicates the number of copies for $i$-th phoneme embedding $\mathbf{x}_i$.
Then, we use a Length Regulator ($\mathcal{LR}$): 
\begin{equation}
    \begin{aligned}
        \mathcal{X}_{mel} = \mathcal{LR}(\mathcal{X}, \mathcal{D}) = \mathcal{LR}(\mathcal{X}, S_d(\mathcal{X})),
     \end{aligned}
     \label{eq:X_mel}
\end{equation}
where $\mathcal{X}_{mel}$ is extended phoneme embedding\footnote{For example, if $d_i=2$, $\mathcal{X}_{mel}$ would be $\mathcal{X}_{mel} = \{..., \mathbf{x}_i, \mathbf{x}_i, ...\}$. For simplicity, we redefine $\mathcal{X}_{mel}$ as $\mathcal{X}_{mel} = \{\mathbf{x}_1, ..., \mathbf{x}_t, ... \mathbf{x}_T\}$} with length $T$. 
To optimise the duration predictor, we use Montreal forced alignment (MFA)~\cite{mcauliffe2017montreal} tools to obtain the ground-truth phoneme duration sequence, and then calculate a mean square error (MSE) loss between ground-truth $\mathcal{\hat{D}}$ and predicted $\mathcal{D}$.
Formally, the loss can be defined as
\begin{equation}
    \begin{aligned}
        \mathcal{L}_{dur} = \frac{1}{L}\sum_{i=1}^{L}(\hat{d}_i - d_i)^2.
     \end{aligned}
     \label{eq:duration_predictor}
\end{equation}

\noindent\textbf{Pitch and Energy Predictors.}
To affect the prosody and volume of speech, following~\cite{ren2020fastspeech}, we employ a pitch predictor $S_p$ and an energy predictor $S_e$, respectively.
Specifically, to predict the pitch contour, we use continuous wavelet transform (CWT) to convert the continuous pitch into pitch spectrogram~\cite{suni2013wavelets,hirose2015speech}, and take it as ground-truth to optimise the pitch predictor $S_p$ by MSE loss:
\begin{equation}
    \begin{aligned}
        \mathcal{L}_{pitch} = \frac{1}{T}\sum_{t=1}^{T}(\hat{p}_t - p_t)^2,
     \end{aligned}
     \label{eq:pitch_predictor}
\end{equation}
where $\hat{p}_t$ and $p_t=S_p(\mathbf{x}_t)$ denote the $t$-th ground-truth and predicted pitch value, respectively.
For energy, we follow the operation in~\cite{ren2020fastspeech} that calculate an L2-norm of the amplitude of each short-time Fourier transform (STFT) frame and take it as energy. The corresponding loss function is
\begin{equation}
    \begin{aligned}
        \mathcal{L}_{energy} = \frac{1}{T}\sum_{t=1}^{T}(\hat{e}_t - e_t)^2,
     \end{aligned}
     \label{eq:enery_predictor}
\end{equation}
where $\hat{e}_t$ and $e_t=S_e(\mathbf{x}_t)$ are the $t$-th ground-truth and predicted energy value, respectively.
Last, we encode each pitch and energy value into the corresponding embedding by the embedding layers $\phi$ and $\varphi$ separately, and then add the pitch and energy embeddings into the extended phoneme embedding $\mathcal{X}_{mel}$. Mathematically, the mel-spectrograms $\mathcal{Y}$ can be generated by
\begin{equation}
    \begin{aligned}
        \mathcal{Y} = g\left(\mathcal{X}_{mel} \oplus \phi(S_p(\mathcal{X}_{mel})) \oplus \varphi(S_e(\mathcal{X}_{mel}))\right),
     \end{aligned}
     \label{eq:Y}
\end{equation}
where $g(\cdot)$ refers to a mel-spectrogram decoder, consisting of 6 FFT blocks~\cite{ren2019fastspeech}.
To optimise the predicted mel-spectrogram, we use the loss function:
\begin{equation}
    \begin{aligned}
        \mathcal{L}_{mel} = \frac{1}{T}\sum_{t=1}^{T}\|\hat{\mathbf{y}}_t - \mathbf{y}_t\|,
     \end{aligned}
     \label{eq:mel_predictor}
\end{equation}
where $\hat{\mathbf{y}}_t$ denotes the $t$-th frame of the ground-truth mel-spectrogram while $\mathbf{y}_t \in \mathcal{Y}$ is the predicted one.

\subsection{Vocoder for Speech Synthesis}
\label{sec:vocoder}

In Figure~\ref{fig:overall_architecture}, to convert the generated mel-spectrogram $\mathcal{Y}$ into time-domain waveform $\mathcal{Y}_{wave}$, we use HiFi-GAN~\cite{kong2020hifi} as our vocoder, which mainly focuses on the raw waveform generation from mel-spectrogram by GANs~\cite{goodfellow2014generative}.
The generator of HiFi-GAN can be divided into two major modules: a transposed convolution (ConvTranspose) network and a multi-receptive field fusion (MRF) module. Specifically, we first upsample the mel-spectrogram $\mathcal{Y}$ by ConvTranspose, which seeks to take an alignment between the length of the output features and the temporal resolution of raw waveforms.
Then, we feed the upsampled features into the MRF module, which consists of multiple residual blocks~\cite{he2016deep}, and take the sum of outputs from these blocks as our predicted waveform. 
Here, we follow the settings of~\cite{kong2020hifi} that use the residual blocks with different kernel sizes and dilation rates to ensure different receptive fields.
%
%
We optimise the vocoder via the objective function that contains an LSGAN-based loss~\cite{mao2017least}, a mel-spectrogram loss~\cite{isola2017image}, and a feature matching loss~\cite{kumar2019melgan}.
In practice, we use the vocoder (\ie, HiFi-GAN) pretrained on LibriSpeech dataset~\cite{panayotov2015librispeech}.

\section{Experiments}\label{sec:experiment}

\begin{table*}[t]
  \centering
  \resizebox{1.0\linewidth}{!}
  {
    \begin{tabular}{c|ccc|cc|cc}
    \toprule
    Methods & MCD $\downarrow$ & MCD-DTW $\downarrow$ & MCD-DTW-SL $\downarrow$ & Id. Acc. $\uparrow$ & Emo. Acc. $\uparrow$ & MOS-naturalness $\uparrow$ & MOS-similarity $\uparrow$ \\
    \midrule
    Ground Truth  &   00.00   &   00.00   & 00.00 & 90.62  & 84.38  & 4.61 $\pm$ 0.15  & 4.74 $\pm$ 0.12 \\
    \midrule
    SV2TTS~\cite{Jia2018TransferLF} &   21.08    &   12.87    & 49.56 & 33.62 & 37.19 &   2.03 $\pm$ 0.22 & 1.92 $\pm$ 0.15\\
    SV2TTS*~\cite{Jia2018TransferLF} &   17.41    &   11.16   & 15.92 & 38.21 & 41.24 &  3.20 $\pm$ 0.20  & 3.09 $\pm$ 0.33 \\
    FastSpeech2~\cite{ren2020fastspeech} &   12.08   &  10.29   & 10.31 & 59.38 & 53.13 &  3.86 $\pm$ 0.07  & 3.75 $\pm$ 0.06  \\
    \midrule
    V2C-Net (Ours) &  \textbf{11.79}    &  \textbf{10.09}   &  \textbf{10.05}  &  \textbf{62.50}  &  \textbf{56.25}  &  \textbf{3.97 $\pm$ 0.06}  & \textbf{3.90 $\pm$ 0.06}  \\
    \bottomrule
    \end{tabular}%
    }
  \caption{Comparison with the state-of-the-art methods. We provide the results of both objective (\ie, MCD, MCD-DTW and MCD-DTW-SL) and subjective evaluation metrics (\ie, MOS-naturalness and MOS-similarity). ``Id. Acc.'' and ``Emo. Acc.'' are the identity and emotion accuracy of the generated speech, respectively. The method with ``*'' refers to a variant taking video (emotion) embedding as an additional input. ``Ground Truth'' denotes the results on ground-truth samples. $\uparrow(\downarrow)$ means that the higher (lower) value is better.}
  \label{tab:closeset}%
\end{table*}%

We evaluate the quality of generated speech in terms of three aspects: 1) objective evaluation, 2) subjective evaluation, and 3) identity and emotion accuracy. The objective and subjective evaluation metrics aim to assess the quality of generated speeches by comparing with ground-truth ones. By contrast, the identity accuracy and emotion accuracy focus on whether the generated speeches involve the desired voice (\ie, identity) and emotion, respectively.
We provide both quantitative and qualitative results on the V2C-Animation dataset. More details are in the following.

\subsection{Evaluation Metrics}

\noindent\textbf{Objective Evaluation Metric.}
To assess the quality of generated speech, we use Mel Cepstral Distortion (MCD)~\cite{kubichek1993mel} metric, which compares Mel Frequency Cepstral Coefficient (MFCC) vectors $\mathcal{C} = \{\mathbf{c}_1, \mathbf{c}_2, ..., \mathbf{c}_i, ..., \mathbf{c}_M\}$ and
$\mathcal{C}' = \{\mathbf{c}'_1, \mathbf{c}'_2, ..., \mathbf{c}'_j, ..., \mathbf{c}'_N\}$ derived from the generated speech and ground truth, respectively. We sum the Euclidean distance over the first $K$ MFCC values:
\begin{equation}\label{eq:MCD}
    \small
    \begin{aligned}
        \mathrm{MCD}(\mathcal{C}, \mathcal{C}') = \frac{1}{T} \sum_{t=1}^{T}d(\mathbf{c}_t, \mathbf{c}'_t) = \frac{1}{T} \sum_{t=1}^{T}\sqrt{\sum_{k=1}^K (c_{t,k}-c'_{t,k})^2},
     \end{aligned}
\end{equation}
where $M=N=T$ refers to the number of speech/audio frames. The $c_{t,k}$ and $c'_{t,k}$ denote the $k$-th MFCC value of the $t$-th speech frame from generated and ground-truth speeches, respectively, while $\mathbf{c}_t = (c_{t,1}, c_{t,2} ..., c_{t,K})$ and $\mathbf{c}'_t = (c'_{t,1}, c'_{t,2} ..., c'_{t,K})$.

Note that the MCD metric requires the lengths of two input speeches to be the same (\ie, $M=N$). When $M\neq N$, the existing voice cloning methods like~\cite{skerry2018towards} simply extend the shorter speech to the length of longer one by padding 0 for the time-domain waveform.
In this way, the value of MCD may become extremely large if the mismatching occurs at the beginning of two speeches. To avoid this issue,
Battenberg \etal\cite{battenberg2020location} use an improved MCD metric, called MCD-DTW, which adopts the Dynamic Time Warping (DTW)~\cite{muller2007dynamic} algorithm to find the minimum MCD between two speeches. 
However, MCD-DTW would achieve a better value as long as there is a match between two speeches. 
This is not reasonable as a better generated speech should have a similar length with the ground truth.

To alleviate the above issues,
we propose a MCD-DTW weighted by Speech Length (MCD-DTW-SL), which evaluates both the length and the quality of alignment between two speeches. 
In MCD-DTW-SL, to evaluate whether the two speeches (\ie, $\mathcal{C}$ and $\mathcal{C}'$) are aligned, we still use DTW algorithm to calculate the minimum distance among them.
Specifically, we compute the cumulative distance $\gamma_{i,j}=d(\mathbf{c}_i, \mathbf{c}'_j)+\text{min}(\gamma_{i-1, j-1}, \gamma_{i-1,j}, \gamma_{i, j-1})$, where $\gamma_{i,j}$ is the minimum cumulative distance from index $(1,1)$ to $(i,j)$. Then, we obtain the objective minimum distance $\gamma_{M,N}$ by accumulating $R$ distances in total, where $\text{max}(M,N)\leq R < M+N-1$.
Besides, considering the influence of lengths, we design a simple but effective coefficient
$\eta=\frac{\text{max}(M,N)}{\text{min}(M,N)}$.
Formally, we calculate the metric
\begin{equation}\label{eq:adv_MCD_DTW}
        \text{MCD-DTW-SL} (\mathcal{C}, \mathcal{C}') =  \frac{\eta}{R} \cdot \gamma_{M,N}.
\end{equation}

\noindent\textbf{Subjective Evaluation Metric.}
To further evaluate the quality of generated speech, we conduct a human study by using a subjective evaluation metric. Specifically, following the settings in~\cite{Jia2018TransferLF}, we use a Mean Opinion Score (MOS) evaluation approach based on subjective listening tests. In this approach, we use the Absolute Category Rating (ACR) scale~\cite{rec1996p} with rating scores from 1 to 5 (\ie, from ``Bad'' to ``Excellent'') in 0.5 point increments. Based on such an approach, we mainly evaluate the generated speeches with respect to naturalness and similarity.
1) \textit{MOS-naturalness}: to assess the naturalness of the generated speech,
we randomly sample 100 generated audios from the testing set and divide them into 4 groups. Each group is rated by a single rater.
%
2) \textit{MOS-similarity}: to evaluate whether the generated speech is well aligned with the desired voice and prosody, we compare each generated speech with the ground-truth one from the same speaker. We use the same samples as when evaluating MOS-naturalness above.
Each pair is rated by the rater according to the similarity between two speeches.

\noindent\textbf{Identity and Emotion Accuracy.}
To evaluate whether the generated speech carries proper speaker identity and emotion, we propose an identity accuracy and an emotion accuracy, respectively. The identity accuracy aims to verify whether the generated speech can be recognised as the same speaker as the input reference audio. Similarly, the emotion accuracy reflects whether the generated speech contains the same emotion like the reference video.
To this end, we first use GE2E~\cite{wan2018generalized} model as our audio encoder
to obtain a fixed-dimensional audio embedding for each speech. Then, based on the audio embeddings $\{\mathbf{a}_{i1}, ..., \mathbf{a}_{iH}\}$\footnote{The embeddings have been normalised using L2 norm.} of the $i$-th speaker, we obtain the centroid $\mathbf{a}'_i$ of the $i$-th speaker by $\mathbf{a}'_i = \mathbb{E}_d [\mathbf{a}_{ih}] = \frac{1}{H} \sum_{d=1}^{H} \mathbf{a}_{ih}$,
where $H$ refers to the number of audio belonging to the $i$-th speaker.
%
Last, we compare the cosine similarity between the embedding of generated speech and each centroid, and then classify it into the category the most similar centroid belongs to.
%
Similarly, we can calculate the emotion accuracy in the same way as well.

\subsection{Quantitative Evaluation}

To evaluate the performance of our method, we compare V2C-Net with several state-of-the-art methods.
In Table~\ref{tab:closeset}, our V2C-Net consistently outperform the existing VC models (\eg, the MCD-DTW-SL of our V2C-Net is $10.05$ while the SV2TTS model only achieves $49.56$).
Besides, we propose a variant of SV2TTS model, called SV2TTS*, which takes the video embedding derived from our method as an additional input. From Table~\ref{tab:closeset}, SV2TTS* achieves better performance compared with SV2TTS, which further demonstrates the effectiveness of our video component. 
Note that the Id.~Acc. and Emo.~Acc. on the ground truth are not 100\%. This is because our pre-trained identity and emotion classification models are not perfect. 

%


\subsection{Qualitative Evaluation}

\begin{figure*}[t]
    \centering
    \includegraphics[width=1.0\linewidth]{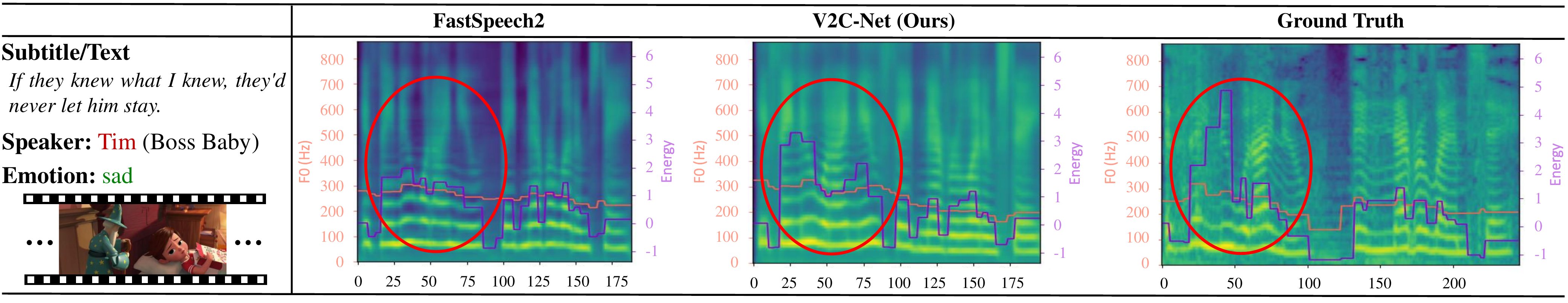}
    \caption{Mel-spectrogram of generated and ground-truth audios. Orange curves are $F_0$ contours, where $F_0$ is fundamental frequency of audio. Purple curves refer to energy (volume) of audio. Horizontal axis is duration of audio.
    We highlight main difference via red circle.
    }
    \label{fig:qualitative evaluation}
\end{figure*}

To further test the performance of generated speech, we show the visualised results of the proposed methods, baseline method and ground-truth, respectively. 
In Figure~\ref{fig:qualitative evaluation}, compared with FastSpeech2, both the energy (volume) curve and the fundamental frequency curve (\ie, $F_0$ curve) of the mel-spectrogram generated by our V2C-Net is more similar to the ground-truth (GT) ones.
Notably, as the duration of audio should be predicted as well, the lengths of the generated audio and the GT one may be different.
More visual results can be found in the supplementary.
%
%

\subsection{Effect of Reference Audio and Video}

To investigate the effect of the reference audio and video, we conduct an ablation study to compare the generated speeches by removing them alternately and show the quantitative results (\ie, identity and emotion accuracy mentioned above) in Table~\ref{tab:ablation}. 
The results show that with the control of reference audio, our V2C-Net achieves higher identity accuracy obviously than the counterpart without reference audio (\ie, from $25.00\%$ to $59.38\%$). After further incorporating the information of reference video, the model obtains the best performance on both two metrics.

\begin{table}[t]
  \centering
  \resizebox{0.80\linewidth}{!}
  {
    \begin{tabular}{c|cc|cc}
    \toprule
          & ref.~A & ref.~V & Id.~Acc. $\uparrow$ & Emo.~Acc. $\uparrow$ \\
    \midrule
    \multirow{2}[4]{*}{V2C-Net} 
          &     &    $\surd$   &    25.00   &  47.61 \\
          &   $\surd$     &       &   59.38    &  53.13  \\
          & $\surd$      & $\surd$      &     \textbf{62.50}  &  \textbf{56.25}  \\
    \bottomrule
    \end{tabular}%
    }
  \caption{Effect of reference audio and video. ``ref.~A'' and ``ref.~V'' denote the reference audio and reference video separately. ``Id.~Acc.'' and ``Emo.~Acc.'' are identity and emotion accuracy of the generated speech, respectively. $\uparrow$ means higher value is better.}
  \label{tab:ablation}%
\end{table}%

\subsection{Comparing Difficulties of V2C and VC Tasks}

To investigate whether our V2C task is more challenging than the conventional VC task, we compare results of the SV2TTS method~\cite{Jia2018TransferLF} on both our V2C-Animation dataset and the existing VC datasets (\eg, VCTK~\cite{yamagishi2019cstr}, LibriSpeech~\cite{panayotov2015librispeech} and Multi-speaker~\cite{skerry2018towards}).
In Table~\ref{tab:difficulty_of_task}, SV2TTS obtains $4.07\pm0.06$ and $3.89\pm0.06$ MOS-naturalness on VCTK and LibriSpeech, respectively, which are higher than that on our V2C-Animation dataset ($2.03\pm0.22$ in Table~\ref{tab:difficulty_of_task}).
Besides, the SV2TTS model achieves $12.37$ MCD value on Multi-speaker dataset~\cite{skerry2018towards}, which is better than the results of the same model on our V2C-Animation dataset (\ie, $21.08$ in Table~\ref{tab:difficulty_of_task}). It demonstrates that the proposed V2C task is more non-trivial as the same VC model obtains worse results on our V2C-Animation dataset than others.

\begin{table}[t]
  \centering
  \resizebox{0.95\linewidth}{!}
  {
    \begin{tabular}{l|l|cc}
    \toprule
    Task  & Dataset & \multicolumn{1}{l}{MCD $\downarrow$} & \multicolumn{1}{l}{MOS-naturalness $\uparrow$} \\
    \midrule
    \multirow{3}[1]{*}{VC} & VCTK~\cite{yamagishi2019cstr}  &   -    & 4.07 $\pm$ 0.06 \\
          & LibriSpeech~\cite{panayotov2015librispeech} &   -    & 3.98 $\pm$ 0.06 \\
          & Multi-speaker~\cite{skerry2018towards} & 12.37 & - \\
    \midrule
    V2C   & V2C-Animation (Ours) & 21.08 & 2.03 $\pm$ 0.22 \\
    \bottomrule
    \end{tabular}%
    }
  \caption{Comparisons on the difficulties of the conventional Voice Cloning (VC) and Visual Voice Cloning (V2C) tasks. We show the performance of SV2TTS~\cite{Jia2018TransferLF} trained on different datasets. VCTK, LibriSpeech and Multi-speaker are widely used in VC task. $\uparrow(\downarrow)$ means higher (lower) value refers to better performance.}
  \label{tab:difficulty_of_task}%
\end{table}%

\subsection{Future Work and Discussion on Social Impacts}\label{sec:future_direction}

\noindent\textbf{Limitations and Future Work.}
In the future, we may extend V2C-Net in two aspects.
First, to grasp the emotion from video, we simply use I3D model as emotion encoder. However, it may not well disentangle emotion from character identity (see results in Table~\ref{tab:ablation}). To alleviate this issue, we may design an emotion-aware loss to capture more discriminative emotion features.
Second, we integrate the multi-modal features (\ie, text, audio, video) by a simple operation (\ie, element-wise add), which may result in sub-optimal performance. Thus, a more promising model for feature fusion is necessary, \eg, COOT~\cite{ging2020coot} or VATT~\cite{akbari2021vatt}. 

\noindent\textbf{Discussion on Social Impacts.}
We provide a new V2C task, which takes advantage of many real-world applications, \eg, movie dubbing or restoring the ability to communicate naturally for users who have lost their voice. However, the technology has a risk to be used maliciously for fake voice generation, which may be misused to financial scams by combining with video deepfakes. To avoid this issue, in this paper, we only focus on the voice generation based on animated movies without any personally identifiable information, \eg, face of the real person.







\section{Conclusion}

In this paper, we propose a novel task, Visual Voice Cloning (V2C), extended from conventional Voice Cloning. 
It seeks to convert a paragraph of text to speech with the desired voice and emotion from reference audio and video, respectively. 
To facilitate the research of this new task, we collect the first V2C-Animation dataset. 
We also design a V2C baseline method, namely Visual Voice Cloning Network (V2C-Net), based on FastSpeech2 (a widely used TTS framework). 
Moreover, to assess the quality of the generated speech, we propose a variant of MCD-DTW, called MCD-DTW-SL, which is weighted by speech length. 
The experimental results demonstrate the effectiveness of our V2C-Net, but it is still far from saturation.


{\small
\bibliographystyle{ieee_fullname}
\bibliography{egbib}

\begin{thebibliography}{10}\itemsep=-1pt

\bibitem{akbari2021vatt}
Hassan Akbari, Linagzhe Yuan, Rui Qian, Wei-Hong Chuang, Shih-Fu Chang, Yin
  Cui, and Boqing Gong.
\newblock Vatt: Transformers for multimodal self-supervised learning from raw
  video, audio and text.
\newblock {\em arXiv preprint arXiv:2104.11178}, 2021.

\bibitem{Arik2018NeuralVC}
Sercan~{\"O}. Arik, Jitong Chen, Kainan Peng, Wei Ping, and Yanqi Zhou.
\newblock Neural voice cloning with a few samples.
\newblock In {\em NeurIPS}, 2018.

\bibitem{arik2017deep}
Sercan~{\"O} Ar{\i}k, Mike Chrzanowski, Adam Coates, Gregory Diamos, Andrew
  Gibiansky, Yongguo Kang, Xian Li, John Miller, Andrew Ng, Jonathan Raiman,
  et~al.
\newblock Deep voice: Real-time neural text-to-speech.
\newblock In {\em ICML}, pages 195--204, 2017.

\bibitem{battenberg2020location}
Eric Battenberg, RJ Skerry-Ryan, Soroosh Mariooryad, Daisy Stanton, David Kao,
  Matt Shannon, and Tom Bagby.
\newblock Location-relative attention mechanisms for robust long-form speech
  synthesis.
\newblock In {\em ICASSP}, pages 6194--6198, 2020.

\bibitem{bian2019multi}
Yanyao Bian, Changbin Chen, Yongguo Kang, and Zhenglin Pan.
\newblock Multi-reference tacotron by intercross training for style
  disentangling, transfer and control in speech synthesis.
\newblock In {\em International Speech Communication Association}, 2019.

\bibitem{blaauw2019data}
Merlijn Blaauw, Jordi Bonada, and Ryunosuke Daido.
\newblock Data efficient voice cloning for neural singing synthesis.
\newblock In {\em ICASSP}, pages 6840--6844, 2019.

\bibitem{carreira2017quo}
Joao Carreira and Andrew Zisserman.
\newblock Quo vadis, action recognition? a new model and the kinetics dataset.
\newblock In {\em CVPR}, pages 6299--6308, 2017.

\bibitem{chen2021adaspeech}
Mingjian Chen, Xu Tan, Bohan Li, Yanqing Liu, Tao Qin, Sheng Zhao, and Tie-Yan
  Liu.
\newblock Adaspeech: Adaptive text to speech for custom voice.
\newblock {\em ICLR}, 2021.

\bibitem{Chen2019SampleEA}
Yutian Chen, Yannis~M. Assael, Brendan Shillingford, D. Budden, Scott~E. Reed,
  H. Zen, Q. Wang, Luis~C. Cobo, Andrew Trask, Ben Laurie, Çaglar
  G{\"u}lçehre, A{\"a}ron van~den Oord, Oriol Vinyals, and N.~D. Freitas.
\newblock Sample efficient adaptive text-to-speech.
\newblock In {\em ICLR}, 2019.

\bibitem{chung2017lip}
Joon~Son Chung, Andrew Senior, Oriol Vinyals, and Andrew Zisserman.
\newblock Lip reading sentences in the wild.
\newblock In {\em CVPR}, pages 3444--3453, 2017.

\bibitem{coppersmith2014dynamic}
Glen Coppersmith and Erin Kelly.
\newblock Dynamic wordclouds and vennclouds for exploratory data analysis.
\newblock In {\em Proceedings of the Workshop on Interactive Language Learning,
  Visualization, and Interfaces}, pages 22--29, 2014.

\bibitem{Gibiansky2017DeepV2}
Andrew Gibiansky, Sercan~{\"O}. Arik, G. Diamos, J. Miller, Kainan Peng, Wei
  Ping, Jonathan Raiman, and Yanqi Zhou.
\newblock Deep voice 2: Multi-speaker neural text-to-speech.
\newblock In {\em NeurIPS}, 2017.

\bibitem{ging2020coot}
Simon Ging, Mohammadreza Zolfaghari, Hamed Pirsiavash, and Thomas Brox.
\newblock Coot: Cooperative hierarchical transformer for video-text
  representation learning.
\newblock {\em NeurIPS}, 2020.

\bibitem{goodfellow2014generative}
Ian Goodfellow, Jean Pouget-Abadie, Mehdi Mirza, Bing Xu, David Warde-Farley,
  Sherjil Ozair, Aaron Courville, and Yoshua Bengio.
\newblock Generative adversarial nets.
\newblock {\em NeurIPS}, 2014.

\bibitem{goodfellow2013challenges}
Ian~J Goodfellow, Dumitru Erhan, Pierre~Luc Carrier, Aaron Courville, Mehdi
  Mirza, Ben Hamner, Will Cukierski, Yichuan Tang, David Thaler, Dong-Hyun Lee,
  et~al.
\newblock Challenges in representation learning: A report on three machine
  learning contests.
\newblock In {\em International Conference on Neural Information Processing},
  pages 117--124, 2013.

\bibitem{he2016deep}
Kaiming He, Xiangyu Zhang, Shaoqing Ren, and Jian Sun.
\newblock Deep residual learning for image recognition.
\newblock In {\em CVPR}, pages 770--778, 2016.

\bibitem{hirose2015speech}
Keikichi Hirose and Jianhua Tao.
\newblock {\em Speech Prosody in Speech Synthesis: Modeling and generation of
  prosody for high quality and flexible speech synthesis}.
\newblock Springer, 2015.

\bibitem{Hsu2019HierarchicalGM}
Wei-Ning Hsu, Y. Zhang, Ron~J. Weiss, H. Zen, Y. Wu, Yuxuan Wang, Yuan Cao, Y.
  Jia, Z. Chen, Jonathan Shen, P. Nguyen, and Ruoming Pang.
\newblock Hierarchical generative modeling for controllable speech synthesis.
\newblock In {\em ICLR}, 2019.

\bibitem{isola2017image}
Phillip Isola, Jun-Yan Zhu, Tinghui Zhou, and Alexei~A Efros.
\newblock Image-to-image translation with conditional adversarial networks.
\newblock In {\em CVPR}, pages 1125--1134, 2017.

\bibitem{ljspeech17}
Keith Ito and Linda Johnson.
\newblock The lj speech dataset.
\newblock \url{https://keithito.com/LJ-Speech-Dataset/}, 2017.

\bibitem{Jia2018TransferLF}
Ye Jia, Y. Zhang, Ron~J. Weiss, Q. Wang, Jonathan Shen, Fei Ren, Z. Chen, P.
  Nguyen, Ruoming Pang, I. Lopez-Moreno, and Y. Wu.
\newblock Transfer learning from speaker verification to multispeaker
  text-to-speech synthesis.
\newblock In {\em NeurIPS}, 2018.

\bibitem{kalchbrenner2018efficient}
Nal Kalchbrenner, Erich Elsen, Karen Simonyan, Seb Noury, Norman Casagrande,
  Edward Lockhart, Florian Stimberg, Aaron Oord, Sander Dieleman, and Koray
  Kavukcuoglu.
\newblock Efficient neural audio synthesis.
\newblock In {\em ICML}, pages 2410--2419, 2018.

\bibitem{Kingma2014AutoEncodingVB}
Diederik~P. Kingma and M. Welling.
\newblock Auto-encoding variational bayes.
\newblock {\em ICLR}, 2014.

\bibitem{kong2020hifi}
Jungil Kong, Jaehyeon Kim, and Jaekyoung Bae.
\newblock Hifi-gan: Generative adversarial networks for efficient and high
  fidelity speech synthesis.
\newblock {\em NeurIPS}, 2020.

\bibitem{kubichek1993mel}
Robert Kubichek.
\newblock Mel-cepstral distance measure for objective speech quality
  assessment.
\newblock In {\em Proceedings of IEEE Pacific Rim Conference on Communications
  Computers and Signal Processing}, 1993.

\bibitem{kumar2019melgan}
Kundan Kumar, Rithesh Kumar, Thibault de Boissiere, Lucas Gestin, Wei~Zhen
  Teoh, Jose Sotelo, Alexandre de Br{\'e}bisson, Yoshua Bengio, and Aaron
  Courville.
\newblock Melgan: Generative adversarial networks for conditional waveform
  synthesis.
\newblock {\em NeurIPS}, 2019.

\bibitem{li2019neural}
Naihan Li, Shujie Liu, Yanqing Liu, Sheng Zhao, and Ming Liu.
\newblock Neural speech synthesis with transformer network.
\newblock In {\em AAAI}, pages 6706--6713, 2019.

\bibitem{mao2017least}
Xudong Mao, Qing Li, Haoran Xie, Raymond~YK Lau, Zhen Wang, and Stephen
  Paul~Smolley.
\newblock Least squares generative adversarial networks.
\newblock In {\em ICCV}, pages 2794--2802, 2017.

\bibitem{mcauliffe2017montreal}
Michael McAuliffe, Michaela Socolof, Sarah Mihuc, Michael Wagner, and Morgan
  Sonderegger.
\newblock Montreal forced aligner: Trainable text-speech alignment using kaldi.
\newblock In {\em Interspeech}, pages 498--502, 2017.

\bibitem{muller2007dynamic}
Meinard M{\"u}ller.
\newblock Dynamic time warping.
\newblock {\em Information Retrieval for Music and Motion}, pages 69--84, 2007.

\bibitem{nachmani2018fitting}
Eliya Nachmani, Adam Polyak, Yaniv Taigman, and Lior Wolf.
\newblock Fitting new speakers based on a short untranscribed sample.
\newblock In {\em ICML}, pages 3683--3691, 2018.

\bibitem{panayotov2015librispeech}
Vassil Panayotov, Guoguo Chen, Daniel Povey, and Sanjeev Khudanpur.
\newblock Librispeech: an asr corpus based on public domain audio books.
\newblock In {\em ICASSP}, 2015.

\bibitem{Ping2018DeepV3}
Wei Ping, Kainan Peng, Andrew Gibiansky, Sercan~{\"O}. Arik, A. Kannan, Sharan
  Narang, Jonathan Raiman, and J. Miller.
\newblock Deep voice 3: Scaling text-to-speech with convolutional sequence
  learning.
\newblock In {\em ICLR}, 2018.

\bibitem{rec1996p}
ITUT Rec.
\newblock P. 800: Methods for subjective determination of transmission quality.
\newblock {\em International Telecommunication Union, Geneva}, 1996.

\bibitem{ren2020fastspeech}
Yi Ren, Chenxu Hu, Xu Tan, Tao Qin, Sheng Zhao, Zhou Zhao, and Tie-Yan Liu.
\newblock Fastspeech 2: Fast and high-quality end-to-end text to speech.
\newblock In {\em ICLR}, 2021.

\bibitem{ren2019fastspeech}
Yi Ren, Yangjun Ruan, X. Tan, Tao Qin, Sheng Zhao, Zhou Zhao, and T. Liu.
\newblock Fastspeech: Fast, robust and controllable text to speech.
\newblock In {\em NeurIPS}, 2019.

\bibitem{rohrbach2015dataset}
Anna Rohrbach, Marcus Rohrbach, Niket Tandon, and Bernt Schiele.
\newblock A dataset for movie description.
\newblock In {\em CVPR}, pages 3202--3212, 2015.

\bibitem{skerry2018towards}
RJ Skerry-Ryan, Eric Battenberg, Ying Xiao, Yuxuan Wang, Daisy Stanton, Joel
  Shor, Ron Weiss, Rob Clark, and Rif~A Saurous.
\newblock Towards end-to-end prosody transfer for expressive speech synthesis
  with tacotron.
\newblock In {\em ICML}, pages 4693--4702, 2018.

\bibitem{stanton2018predicting}
Daisy Stanton, Yuxuan Wang, and RJ Skerry-Ryan.
\newblock Predicting expressive speaking style from text in end-to-end speech
  synthesis.
\newblock In {\em Spoken Language Technology Workshop}, pages 595--602, 2018.

\bibitem{suni2013wavelets}
Antti~Santeri Suni, Daniel Aalto, Tuomo Raitio, Paavo Alku, Martti Vainio,
  et~al.
\newblock Wavelets for intonation modeling in hmm speech synthesis.
\newblock In {\em ISCA Workshop on Speech Synthesis}, 2013.

\bibitem{Taigman2018VoiceLoopVF}
Yaniv Taigman, Lior Wolf, A. Polyak, and Eliya Nachmani.
\newblock Voiceloop: Voice fitting and synthesis via a phonological loop.
\newblock In {\em ICLR}, 2018.

\bibitem{tapaswi2016movieqa}
Makarand Tapaswi, Yukun Zhu, Rainer Stiefelhagen, Antonio Torralba, Raquel
  Urtasun, and Sanja Fidler.
\newblock Movieqa: Understanding stories in movies through question-answering.
\newblock In {\em CVPR}, pages 4631--4640, 2016.

\bibitem{tomar2006converting}
Suramya Tomar.
\newblock Converting video formats with ffmpeg.
\newblock {\em Linux Journal}, page~10, 2006.

\bibitem{Valle2020MellotronME}
Rafael Valle, Jason Li, R. Prenger, and Bryan Catanzaro.
\newblock Mellotron: Multispeaker expressive voice synthesis by conditioning on
  rhythm, pitch and global style tokens.
\newblock In {\em ICASSP}, pages 6189--6193, 2020.

\bibitem{vaswani2017attention}
Ashish Vaswani, Noam Shazeer, Niki Parmar, Jakob Uszkoreit, Llion Jones,
  Aidan~N Gomez, {\L}ukasz Kaiser, and Illia Polosukhin.
\newblock Attention is all you need.
\newblock In {\em NeurIPS}, pages 5998--6008, 2017.

\bibitem{wan2018generalized}
Li Wan, Quan Wang, Alan Papir, and Ignacio~Lopez Moreno.
\newblock Generalized end-to-end loss for speaker verification.
\newblock In {\em ICASSP}, pages 4879--4883, 2018.

\bibitem{Wang2017TacotronTE}
Yuxuan Wang, R. Skerry-Ryan, Daisy Stanton, Y. Wu, Ron~J. Weiss, Navdeep
  Jaitly, Z. Yang, Y. Xiao, Z. Chen, S. Bengio, Quoc~V. Le, Yannis
  Agiomyrgiannakis, R. Clark, and R. Saurous.
\newblock Tacotron: Towards end-to-end speech synthesis.
\newblock In {\em International Speech Communication Association}, 2017.

\bibitem{wang2018style}
Yuxuan Wang, Daisy Stanton, Yu Zhang, RJ-Skerry Ryan, Eric Battenberg, Joel
  Shor, Ying Xiao, Ye Jia, Fei Ren, and Rif~A Saurous.
\newblock Style tokens: Unsupervised style modeling, control and transfer in
  end-to-end speech synthesis.
\newblock In {\em ICML}, pages 5180--5189, 2018.

\bibitem{Whitehill2020MultiReferenceNT}
M. Whitehill, S. Ma, Daniel McDuff, and Yale Song.
\newblock Multi-reference neural tts stylization with adversarial cycle
  consistency.
\newblock In {\em International Speech Communication Association}, 2020.

\bibitem{yamagishi2019cstr}
Junichi Yamagishi, Christophe Veaux, Kirsten MacDonald, et~al.
\newblock Cstr vctk corpus: English multi-speaker corpus for cstr voice cloning
  toolkit.
\newblock 2019.

\bibitem{yan2021adaspeech}
Yuzi Yan, Xu Tan, Bohan Li, Tao Qin, Sheng Zhao, Yuan Shen, and Tie-Yan Liu.
\newblock Adaspeech 2: Adaptive text to speech with untranscribed data.
\newblock In {\em ICASSP}, pages 6613--6617, 2021.

\bibitem{zen2019libritts}
Heiga Zen, Viet Dang, Rob Clark, Yu Zhang, Ron~J Weiss, Ye Jia, Zhifeng Chen,
  and Yonghui Wu.
\newblock Libritts: A corpus derived from librispeech for text-to-speech.
\newblock {\em arXiv preprint arXiv:1904.02882}, 2019.

\end{thebibliography}
}


\clearpage

\appendix

We organise the supplementary materials as follows.
\vspace{-5pt}
\begin{itemize}
	\item In Section~\ref{sec:V2C_animation_dataset}, we provide a detailed analysis of the proposed V2C-Animation dataset, \eg, word cloud, distribution of utterance length, \etc.
	\vspace{-8pt}
	\item In Section~\ref{sec:qualitative_results}, we show the qualitative results  (in ``\textit{comparison\_with\_SoTA.mp4}'') of our V2C-Net, baseline, and ground truth. Besides, we also investigate whether the reference audio can control the voice of generated speeches in the proposed V2C-Net (results in ``\textit{voice\_cloning.mp4}'').
	\vspace{-8pt}
	\item In Section~\ref{sec:more_sample_V2C_related_dataset}, we pose more visualised pitch tricks of the samples from the V2C-Animation dataset and the related datasets.
	\vspace{-8pt}
	\item In Section~\ref{sec:more_sample_mel}, we exhibit more visual results of the mel-spectrogram derived from our V2C-Net with comparisons against baseline and ground truth.
	\vspace{-8pt}
	\item In Section~\ref{sec:implementation_details}, we report the implementation details.
	\vspace{-8pt}
	\item In Section~\ref{sec:more_details_vocoder}, we depict more details of the vocoder.
\end{itemize}


\section{More Analysis of V2C-Animation Dataset}\label{sec:V2C_animation_dataset}

\subsection{Word Cloud and Count}\label{subsec:word_cloud}

\begin{figure}[h]
	\centering
	\includegraphics[width=1.0\linewidth]{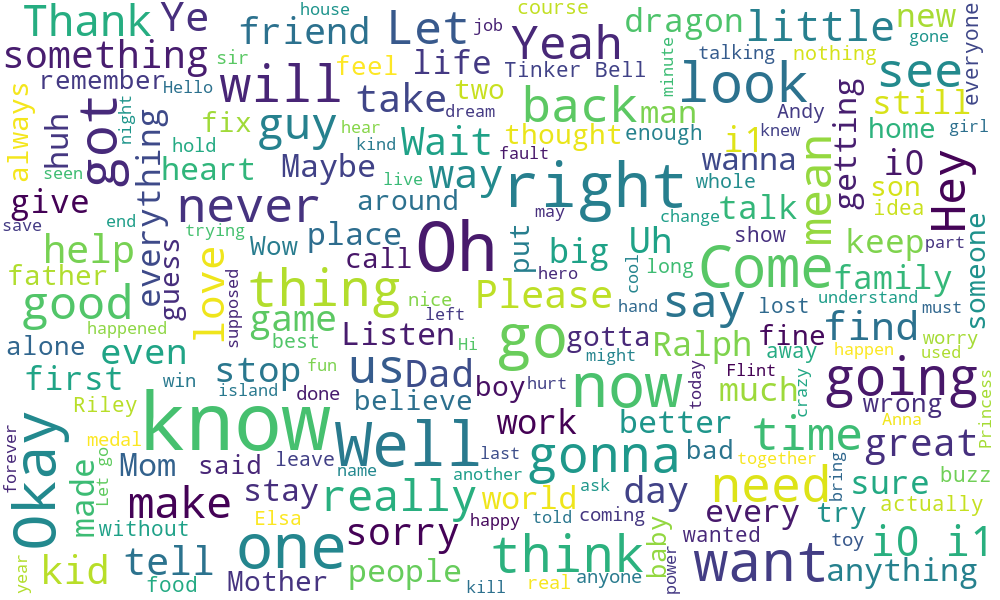}
	\caption{Word cloud of the texts on our V2C-Animation dataset.}
	\label{fig:word_cloud}
\end{figure}

In Figure~\ref{fig:word_cloud}, we visualise the texts/subtitles of our V2C-Animation dataset as Venn-style word cloud~\cite{coppersmith2014dynamic}, where the size of each word refers to the harmonic mean of its count.

\begin{figure}[h]
	\centering
	\includegraphics[width=1.0\linewidth]{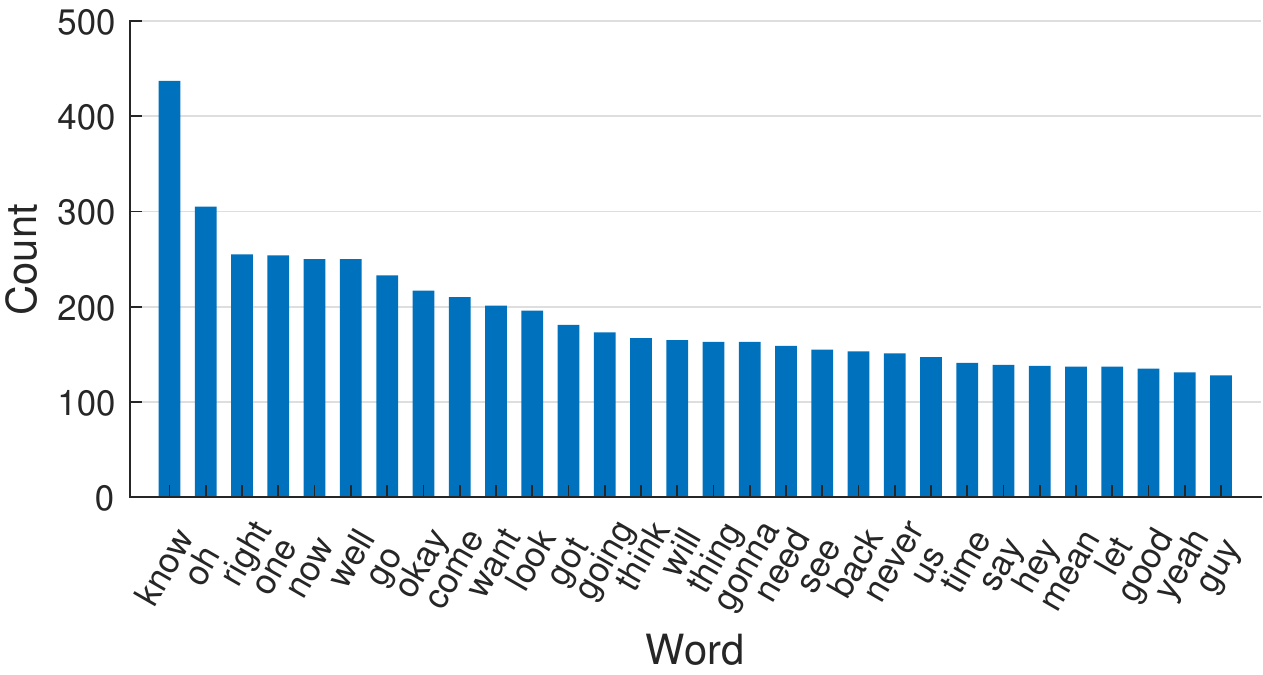}
	\caption{Top 30 words on V2C-Animation along with the counts.}
	\label{fig:word_count}
\end{figure}

Besides, we also provide the top 30 words on our V2C-Animation dataset along with their counts in Figure~\ref{fig:word_count}. More results (top 100) are in the following:\\
(`know', 437), (`oh', 305), (`right', 255), (`one', 254), (`now', 250), (`well', 250), (`go', 233), (`okay', 217), (`come', 210), (`want', 201), (`look', 196), (`got', 181), (`going', 173), (`think', 167), (`will', 165), (`thing', 163), (`gonna', 163), (`need', 159), (`see', 155), (`back', 153), (`never', 151), (`us', 147), (`time', 141), (`say', 139), (`hey', 138), (`mean', 137), (`let', 137), (`good', 135), (`yeah', 131), (`guy', 128), (`really', 124), (`make', 124), (`thank', 124), (`little', 112), (`way', 108), (`love', 108), (`ye', 108), (`find', 104), (`help', 97), (`tell', 96), (`wait', 95), (`take', 93), (`kid', 92), (`please', 91), (`sorry', 88), (`something', 87), (`great', 87), (`dad', 87), (`friend', 84), (`day', 82), (`game', 80), (`stop', 75), (`even', 75), (`Uh', 74), (`big', 67), (`work', 66), (`Ralph', 66), (`much', 62), (`give', 62), (`first', 61), (`everything', 60), (`new', 59), (`still', 58), (`life', 58), (`keep', 58), (`dragon', 58), (`family', 57), (`sure', 56), (`made', 56), (`talk', 55), (`world', 53), (`place', 53), (`heart', 53), (`every', 53), (`maybe', 53), (`stay', 52), (`wanna', 51), (`better', 51), (`people', 50), (`huh', 50), (`anything', 50), (`getting', 49), (`thought', 48), (`man', 48), (`mom', 48), (`listen', 48), (`guess', 47), (`fine', 47), (`around', 47), (`gotta', 46), (`believe', 46), (`two', 45), (`someone', 45), (`home', 45), (`call', 45), (`boy', 45), (`son', 44), (`put', 43), (`fix', 43), (`always', 43)

\subsection{Distribution of Emotion Labels}

\begin{figure}[h]
	\centering
	\includegraphics[width=0.90\linewidth]{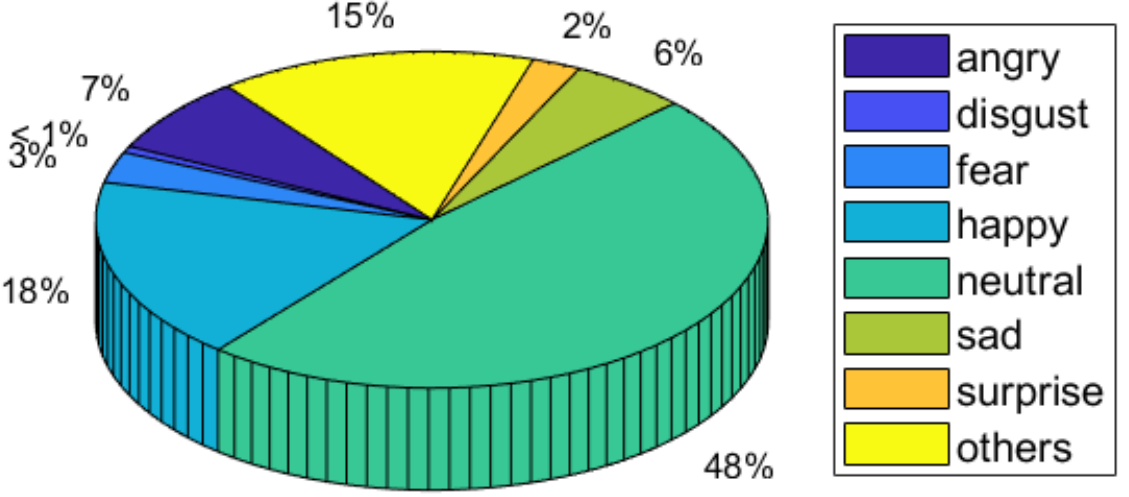}
	\caption{Distribution of emotion labels on V2C-Animation.}
	\label{fig:emotion_distribution}
\end{figure}

Following the categories of FER-2013~\cite{goodfellow2013challenges} (a dataset for human facial expression recognition), we divide the collected video/audio clips into 8 types (\ie, 0: angry, 1: disgust, 2: fear, 3: happy, 4: neutral, 5: sad, 6: surprise, and 7: others).
The number and distribution of each emotion label can be found in Table~\ref{tab:emotion_count} and Figure~\ref{fig:emotion_distribution}, respectively.

\begin{table}[h]
	\centering
	\resizebox{0.68\linewidth}{!}
	{
		\begin{tabular}{c|cccc}
			\toprule
			Emotion & angry & disgust & fear & happy \\
			\midrule
			Count & 756 & 64  & 305 & 1799 \\
			\midrule
			\midrule
			Emotion & neutral & sad & surprise & others \\
			\midrule
			Count & 4919 & 572 & 240 & 1562 \\
			\bottomrule
		\end{tabular}%
	}
	\caption{Counts of the emotion labels on V2V-Animation dataset.}
	\label{tab:emotion_count}%
\end{table}%

\subsection{Distribution of Utterance Length}

\begin{figure}[h]
	\centering
	\includegraphics[width=0.95\linewidth]{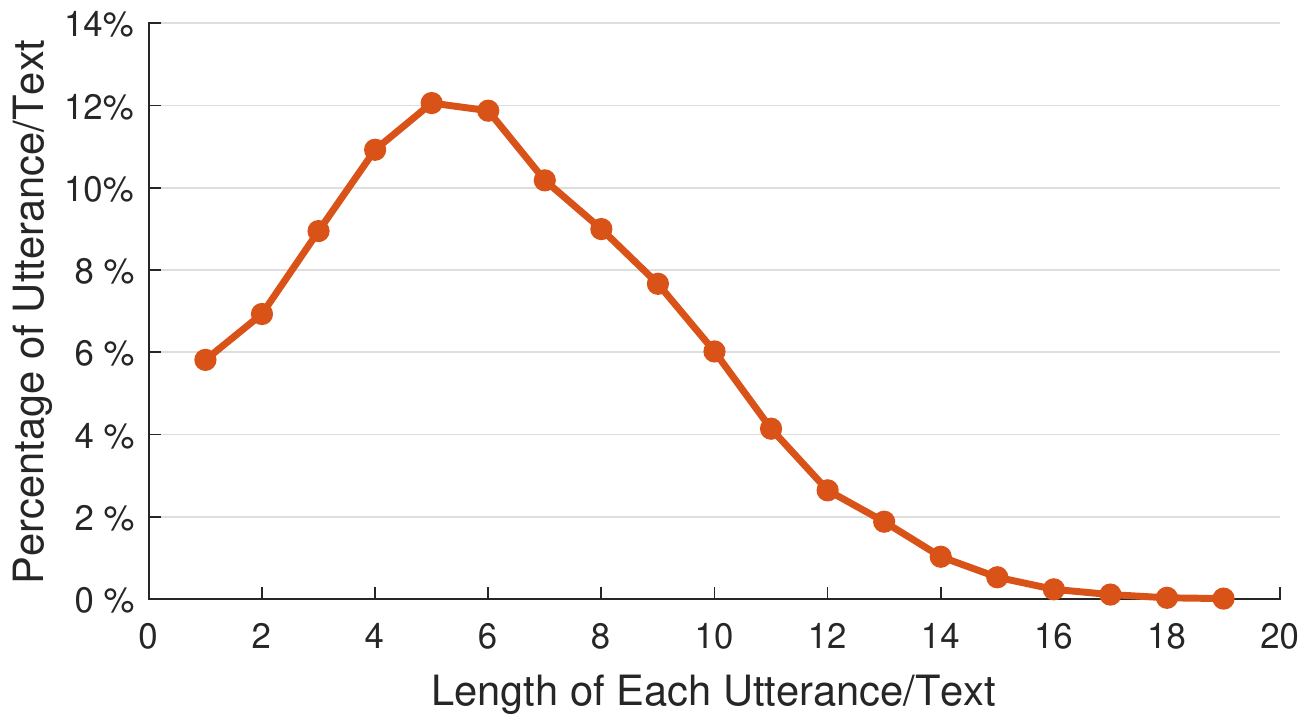}
	\caption{Distribution of utterance/text length.}
	\label{fig:length_distribution}
\end{figure}

Figure~\ref{fig:length_distribution} exhibits the distribution of utterance/text length on V2C-Animation dataset, which shows that most utterance range from 3 to 8 words. Besides, we also list the number of utterance/text and their corresponding percentages in the following (format: \textit{length, count, percentage}):\\
(1, 594, 5.81\%), (2, 708, 6.93\%), (3, 914, 8.95\%), (4, 1116, 10.92\%), (5, 1232, 12.06\%), (6, 1213, 11.87\%), (7, 1040, 10.18\%), (8, 919, 8.99\%), (9, 783, 7.66\%), (10, 615, 6.02\%), (11, 423, 4.14\%), (12, 270, 2.64\%), (13, 192, 1.88\%), (14, 105, 1.03\%), (15, 54, 0.53\%), (16, 24, 0.23\%), (17, 11, 0.11\%), (18, 3, 0.03\%), (19, 1, 0.01\%)

\subsection{More Examples of Subtitle and Video Clip}\label{sec:example_subtitle_video_clip}

\begin{figure}[h]
	\centering
	\includegraphics[width=1.0\linewidth]{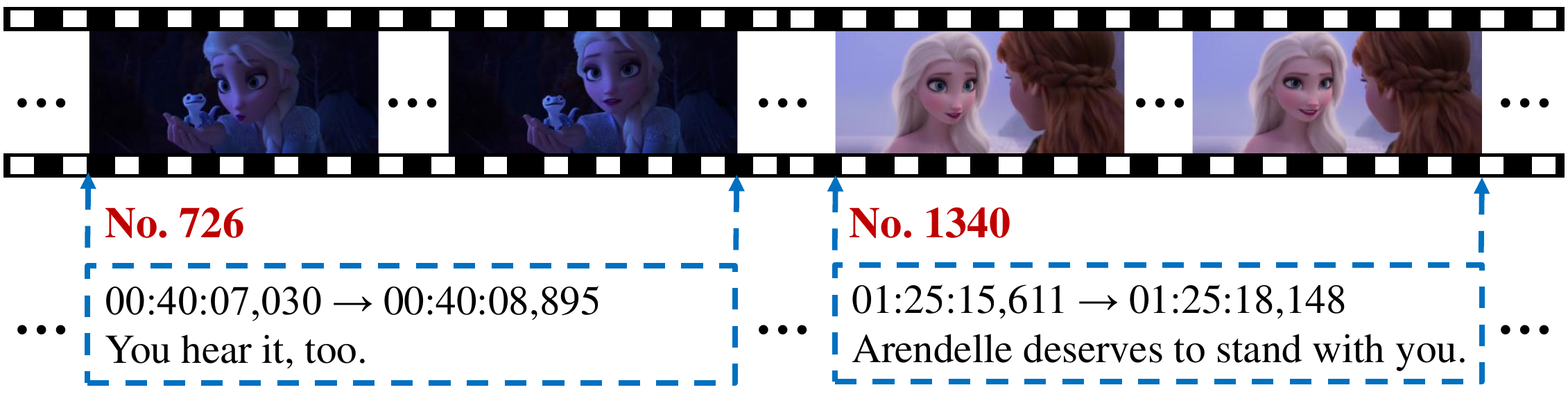}
	\caption{Examples of how to cut a movie into a series of video clips according to subtitle files. Note that the subtitle files contain both starting and ending time-stamps for each video clip.}
	\label{fig:examples_subtitle_crop_video2}
\end{figure}

We show several examples of how to crop movies based on a corresponding subtitle file. Here, we use an SRT type subtitle file. Besides the subtitles/texts, the SRT file also contains starting and ending time-stamps to ensure the subtitles match with video and audio. The sequential number of subtitle (\eg, No.~726 and No.~1340 in Figure~\ref{fig:examples_subtitle_crop_video2}) indicates the index of each video clip. Based on the SRT file, we cut movie into a series of video clips via FFmpeg toolkit~\cite{tomar2006converting} (an automatic audio and video processing toolkit).

\subsection{Samples of Character's Emotion}

\begin{figure}[h]
	\centering
	\includegraphics[width=1.0\linewidth]{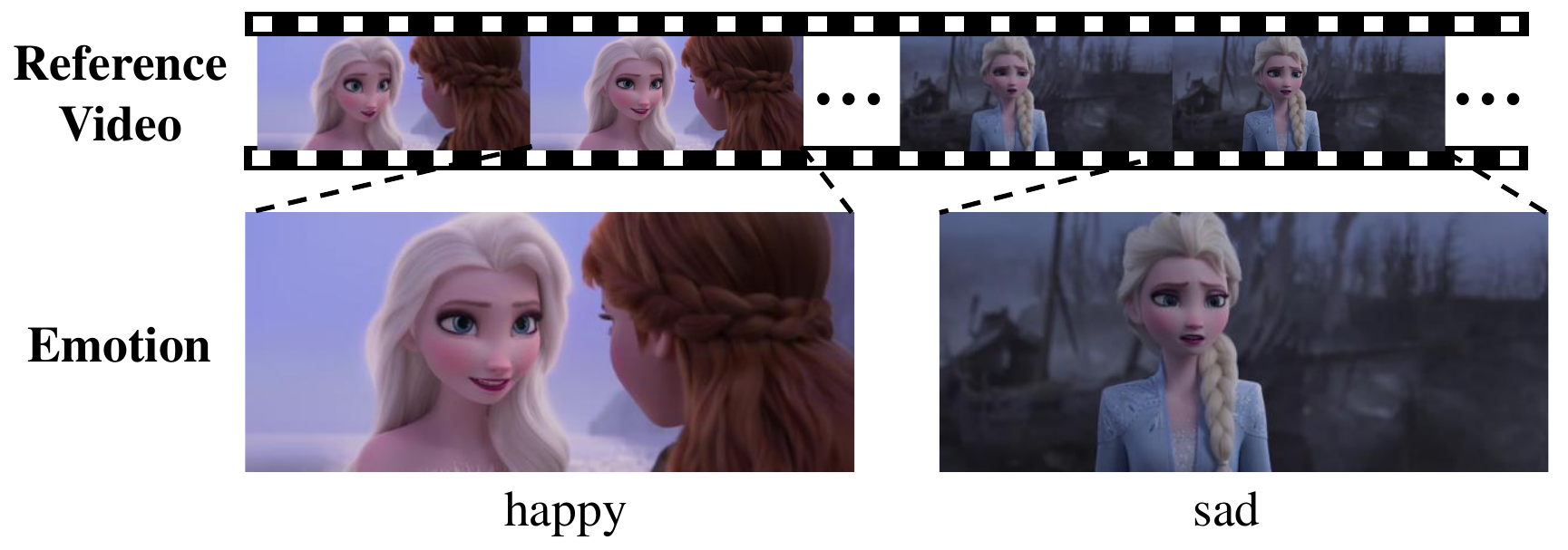}
	\caption{Samples of the character's emotion (\eg, happy and sad) involved in the reference video. Here, we take \textit{Elsa} (a character in movie \textit{Frozen}) as an example.}
	\label{fig:emotion_sample}
\end{figure}

Figure~\ref{fig:emotion_sample} shows some samples of the reference videos on V2C-Animation dataset with their corresponding emotions.

\subsection{List of Animated Movies and Characters}

\begin{figure*}[h]
	\centering
	\includegraphics[width=1.0\linewidth]{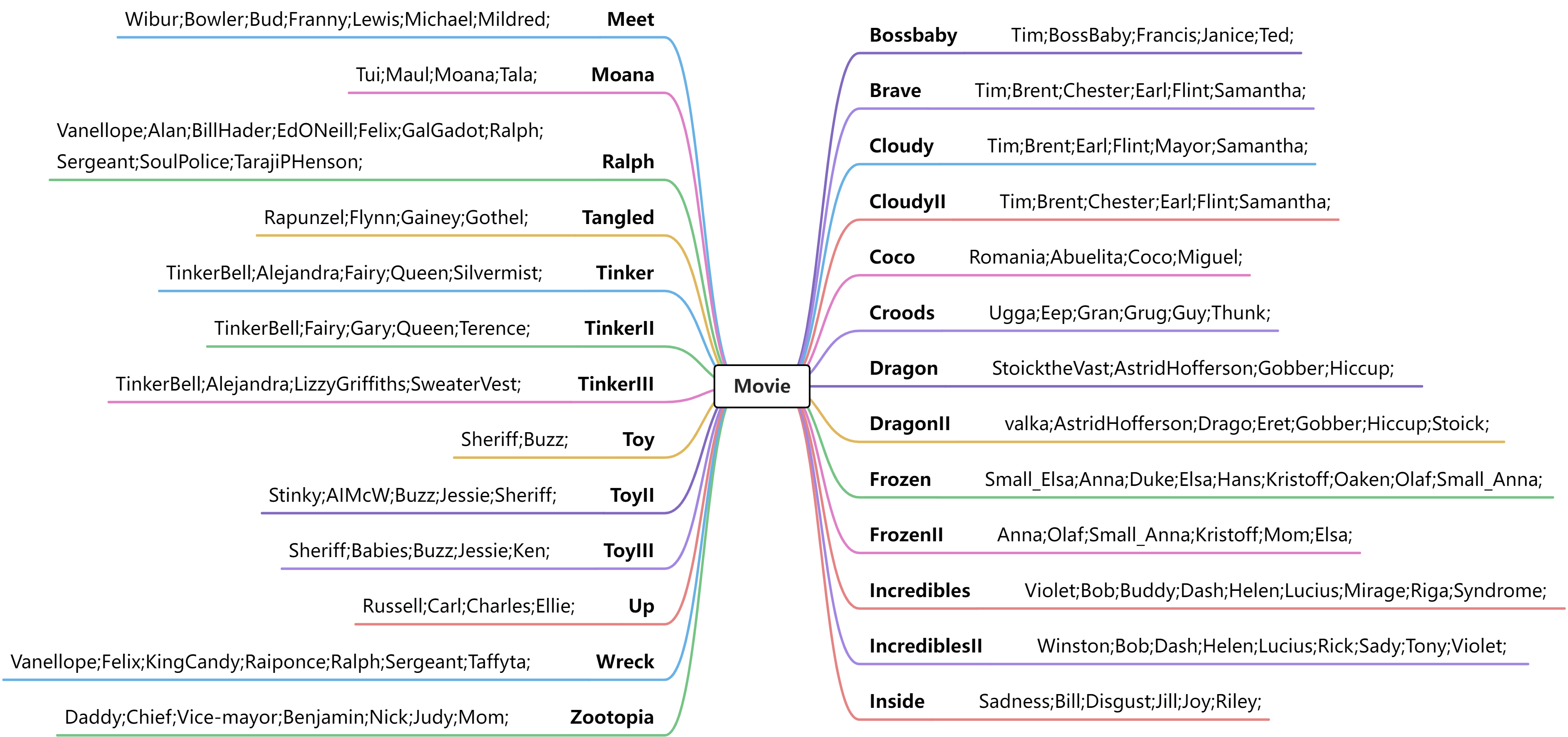}
	\caption{Movies with the corresponding speakers/characters on the V2C-Animation dataset.}
	\label{fig:movie_speeker}
\end{figure*}

As shown in Figure~\ref{fig:movie_speeker}, we report all the names of our collected animated movies with their corresponding characters/speakers on the V2C-Animation dataset.

\section{Analysis of Qualitative Results}\label{sec:qualitative_results}

To further assess the quality of the generated speeches, we provide a video in this supplementary, which compares the generated audios from our V2C-Net, baseline method (\ie, FastSpeech2~\cite{ren2020fastspeech}), and ground truth (\ie, ``\textit{comparison\_with\_SoTA.mp4}'').
Besides, to investigate whether the proposed V2C-Net is able to clone the voice from reference audio, we fix the input text/subtitle and reference video. Then, we generate speeches using the voice derived from different reference audios (\ie, ``\textit{voice\_cloning.mp4}'').

\section{V2C-Animation vs. Related Datasets}\label{sec:more_sample_V2C_related_dataset}

\begin{figure*}[t]
	\centering
	\includegraphics[width=\linewidth]{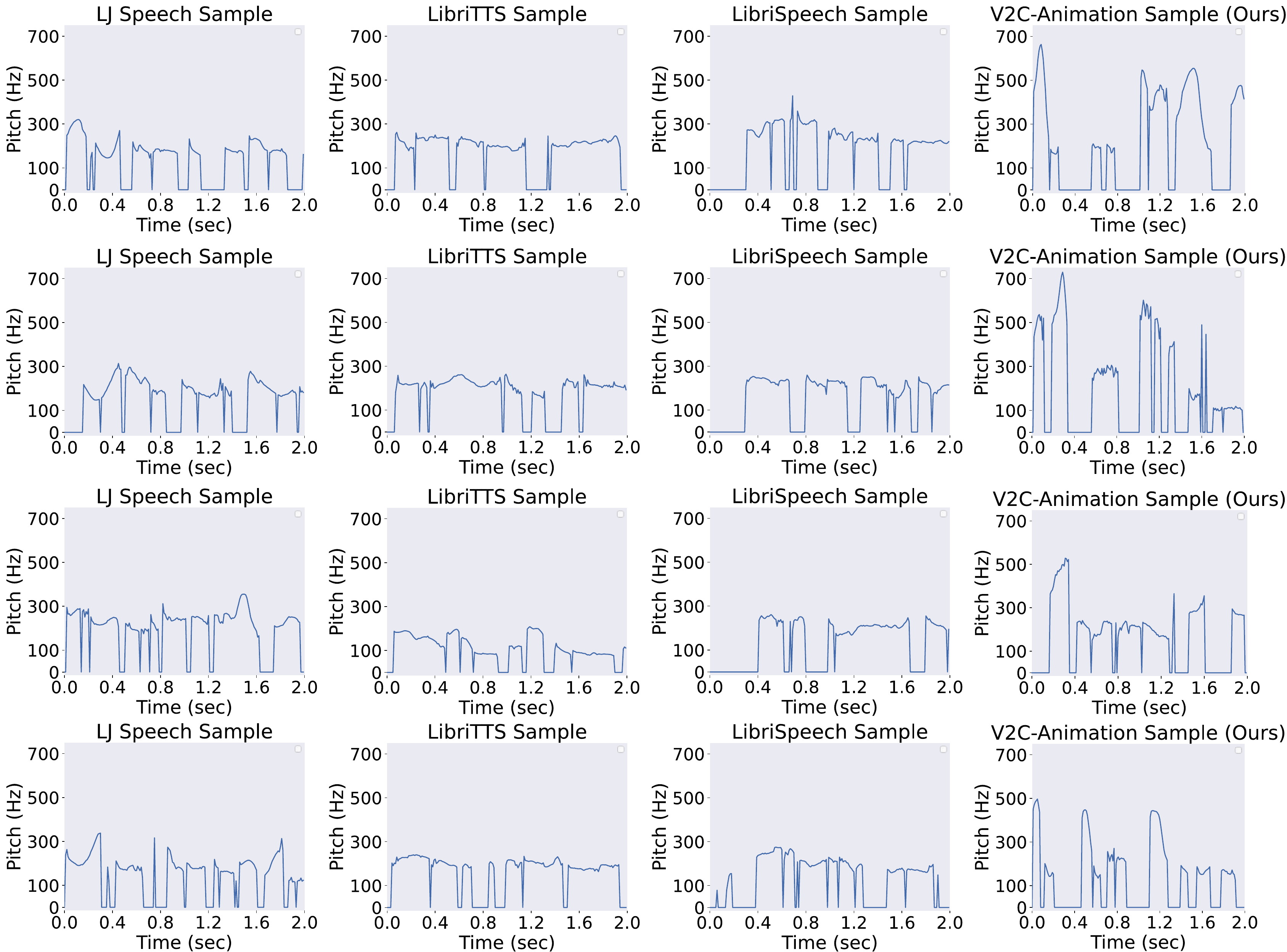}
	\caption{Visual comparison between our V2C-Animation dataset and the related datasets (\ie, LJ Speech, LibriSpeech and LibriTTS). A pitch of 0 Hz refers to an unvoiced segment.}
	\label{fig:MoreSamples_dataset_comparision}
\end{figure*}

To compare the differences between the collected V2C-Animation dataset and several related datasets (\ie, LJ Speech, LibriSpeech and LibriTTS), we visualise the pitch tricks of the samples from our dataset and others. Due to the varying lengths of audios, for a fair comparison, we cut two seconds of audio from each sample. As shown in Figure~\ref{fig:MoreSamples_dataset_comparision}, the audio pitches from the existing datasets are more smooth and their ranges of frequency (Hz) are narrower than ours.
%

\section{More Visual Results of Mel-spectrogram}\label{sec:more_sample_mel}

\begin{figure*}[t]
	\centering
	\includegraphics[width=1.0\linewidth]{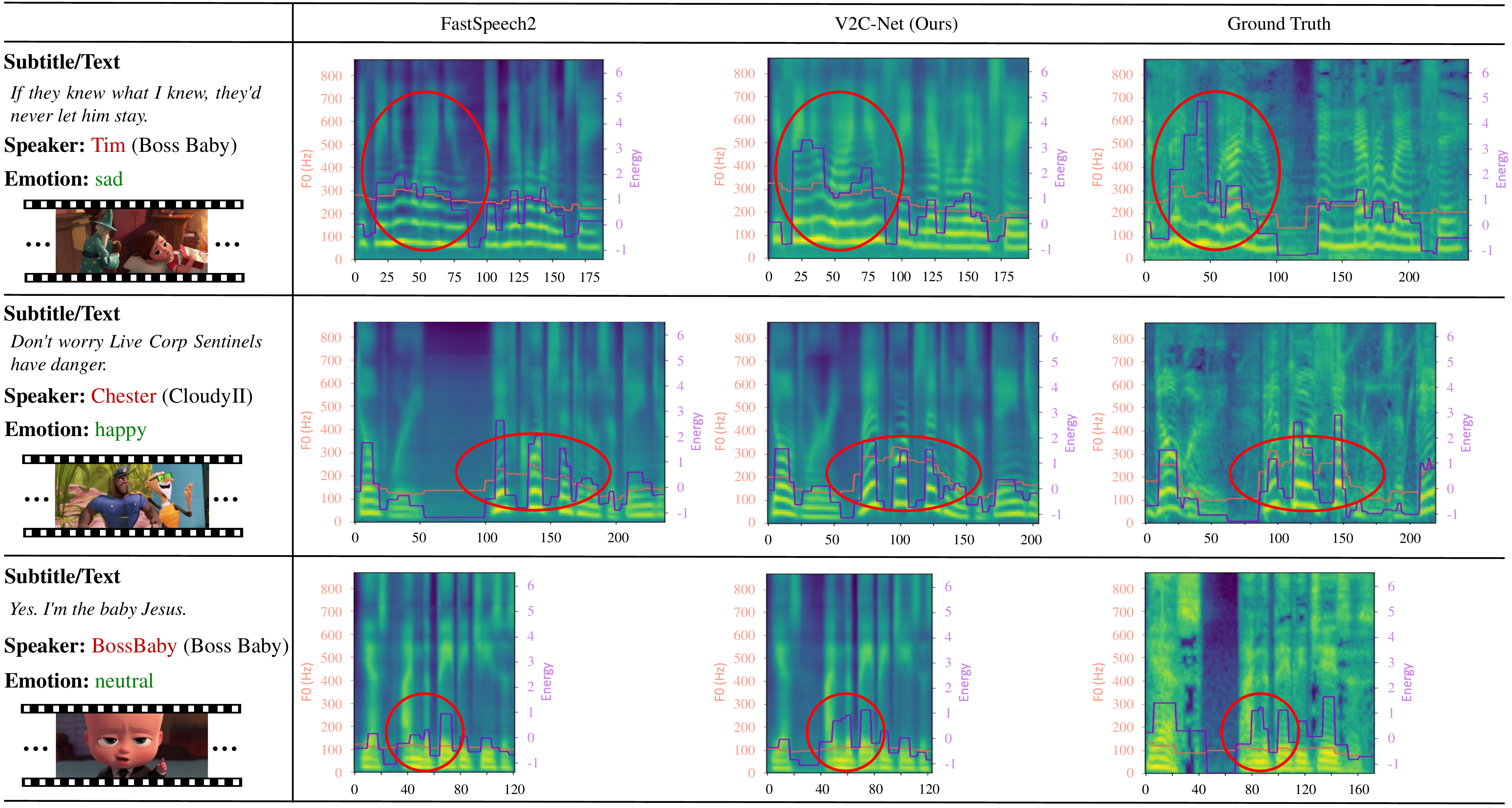}
	\caption{More visualised mel-spectrograms of generated and ground-truth audios. The orange curves are $F_0$ contours, where $F_0$ denotes the fundamental frequency of audio. The purple curves refer to energy (volume) of audio. Horizontal axis is the duration of audio.
		We highlight the main difference via red circle.}
	\label{fig:MoreSamples_visual_mel}
\end{figure*}

We provide more visualised results of our V2C-Net with comparisons against the baseline method and ground truth.
As shown in Figure~\ref{fig:MoreSamples_visual_mel}, the mel-spectrograms generated by the proposed V2C-Net are more similar to the ground-truth ones.
Note that the baseline method FastSpeech2 does not take the reference videos (\ie, emotions) as inputs, which may lead to some misses of the prosody involving in the videos.
The results further demonstrate the effect of the reference video when generating speech with rich emotions.
Besides, the ranges of pitch for the mel-spectrograms are various due to the different emotions. For example, the pitch of the mel-spectrogram would be more drastic  with the emotions ``\textit{happy}'' or ``\textit{sad}'', while it would be more smooth if the emotion is ``\textit{neutral}''.

\section{Implementation Details}\label{sec:implementation_details}

%
For the speaker encoder $f_{spk}$, we use the same architecture as~\cite{wan2018generalized}, comprising three LSTM layers. The audio encoder maps a sequence of mel-spectrogram frames, derived from the reference audio, to a vector with a fixed dimension of 256.
We optimise the model with a generalised end-to-end speaker verification loss, which ensure features from the same speaker are more similar than ones from different speakers.
%
%
For the emotion encoder $f_{emo}$, we use a conventional I3D model~\cite{carreira2017quo}, trained on our V2C-Animation dataset~\cite{carreira2017quo} with $64\times10^3$ iterations and final output a vector with 1024 dimensions.
For our synthesizer, we train the text encoder $f_{txt}$ and the synthesizer in an end-to-end manner with $16$ batch size and $2\times10^6$ iterations on our proposed V2C-Animation dataset.
%
%
We train all models on a single GPU device (GeForce RTX 3090).

\section{Details of Vocoder}\label{sec:more_details_vocoder}

To synthesise the waveform of the speech from our generated mel-spectrogram, we use HiFi-GAN~\cite{kong2020hifi} as our vocoder.
The HiFi-GAN model is based on Generative Adversarial Networks (GANs)~\cite{goodfellow2014generative}, which consists of one generator and two discriminators, \ie, a multi-period discriminator (MPD) and a multi-scale discriminator (MSD).

The generator of HiFi-GAN can be divided into two major modules: a transposed convolution (ConvTranspose) network and a multi-receptive field fusion (MRF) module. Specifically, we first upsample the mel-spectrogram by ConvTranspose, which seeks to take an alignment between the length of the output features and the temporal resolution of raw waveforms.
Then, we feed the upsampled features into MRF module, which consists of multiple residual blocks~\cite{he2016deep}, and take the sum of outputs from these blocks as our predicted waveform. 
Here, the residual blocks with different kernel sizes and dilation rates are used to ensure different receptive field.

For the two discriminators, the multi-period discriminator (MPD) contains several sub-discriminators, where each sub-discriminator handles a specific periodic part of the input audio.
By contrast, the multi-scale discriminator (MSD) proposed in MelGAN~\cite{kumar2019melgan}, consisting of three sub-discriminators, tries to capture the consecutive patterns and long-term dependencies from input audio.

The generator and discriminators are trained adversarially, aiming to improve the training stability and the model performance.
Specifically, the vocoder (\ie, HiFi-GAN) is optimised via the objective function that contains an LSGAN-based loss~\cite{mao2017least}, a mel-spectrogram loss~\cite{isola2017image}, and a feature matching loss~\cite{kumar2019melgan}.
In practice, we use the vocoder (\ie, HiFi-GAN) pretrained on the LibriSpeech dataset~\cite{panayotov2015librispeech}.

\end{document}